\documentclass{article}

\usepackage{microtype}
\usepackage{graphicx}
\usepackage{subfigure}
\usepackage{booktabs} 

\usepackage{comment}
\usepackage{amsmath}
\usepackage{amssymb}
\usepackage{amsfonts}
\usepackage{dsfont}
\usepackage{url}
\usepackage[ruled,lined]{algorithm2e}
\usepackage{multirow}
\usepackage{adjustbox}
\usepackage{amsthm}

\newtheorem{theorem}{Theorem}

\newtheorem{proposition}{Proposition}

\newtheorem*{proof*}{Proof}
\newtheorem{remark}{Remark}

\usepackage{balance}

\usepackage[accepted]{icml2018}

\SetAlgoCaptionSeparator{.}

\newcommand{\eg}{\emph{e.g.}} 
\newcommand{\ie}{\emph{i.e.}} 

\newcommand{\etal}{\emph{et al.}}

\icmltitlerunning{MentorNet: Learning Data-Driven Curriculum for Deep Neural Networks}

\begin{document}

\twocolumn[
\icmltitle{MentorNet: Learning Data-Driven Curriculum \\ for Very Deep Neural Networks
on Corrupted Labels}



\icmlsetsymbol{equal}{*}

\begin{icmlauthorlist}
\icmlauthor{Lu Jiang}{goo}
\icmlauthor{Zhengyuan Zhou}{stan}
\icmlauthor{Thomas Leung}{goo}
\icmlauthor{Li-Jia Li}{goo}
\icmlauthor{Li Fei-Fei}{goo,stan}
\end{icmlauthorlist}

\icmlaffiliation{goo}{Google Inc., Mountain View, United States}
\icmlaffiliation{stan}{Stanford University, Stanford, United States}
\icmlcorrespondingauthor{Lu Jiang}{lujiang@google.com}

\icmlkeywords{Curriculum Learning, Regularization, Very Deep Neural Networks, Noisy Labels}
\vskip 0.3in
]


\printAffiliationsAndNotice{}  

\begin{abstract}
Recent deep networks are capable of memorizing the entire data even when the labels are completely random. To overcome the overfitting on corrupted labels, we propose a novel technique of learning another neural network, called MentorNet, to supervise the training of the base deep networks, namely, StudentNet. During training, MentorNet provides a curriculum (sample weighting scheme) for StudentNet to focus on the sample the label of which is probably correct. Unlike the existing curriculum that is usually predefined by human experts, MentorNet learns a data-driven curriculum dynamically with StudentNet. Experimental results demonstrate that our approach can significantly improve the generalization performance of deep networks trained on corrupted training data. Notably, to the best of our knowledge, we achieve the best-published result on WebVision, a large benchmark containing 2.2 million images of real-world noisy labels. The code are at {\small \url{https://github.com/google/mentornet}}.
\end{abstract}

\vspace{-5mm}
\section{Introduction}
Zhang \etal~\yrcite{zhang2017understanding} found that deep convolutional neural networks (CNNs) are capable of memorizing the entire data even with corrupted labels, where some or all true labels are replaced with random labels. It is a consensus that deeper CNNs usually lead to better performance. However, the ability of deep CNNs to overfit or memorize the corrupted labels can lead to very poor generalization performance~\cite{zhang2017understanding}. Recently, Neyshabur \etal~\yrcite{neyshabur2017exploring} and Arpit \etal~\yrcite{arpit2017closer} proposed deep learning generalization theories to explain this interesting phenomenon.

This paper studies how to overcome the corrupted label for deep CNNs, so as to improve generalization performance on the clean test data. Although learning models on weakly labeled data might not be novel, improving deep CNNs on corrupted labels is clearly an under-studied problem and worthy of exploration, as deep CNNs are more prone to overfitting and memorizing corrupted labels~\cite{zhang2017understanding}. To address this issue, we focus on training very deep CNNs from scratch, such as resnet-101~\cite{he2016deep} or inception-resnet~\cite{szegedy2017inception} which has a few hundred layers and orders-of-magnitude more parameters than the number of training samples. These networks can achieve the state-of-the-art result but perform poorly when trained on corrupted labels.

Inspired by the recent success of Curriculum Learning (CL), this paper tackles this problem using CL~\cite{bengio2009curriculum}, a learning paradigm inspired by the cognitive process of human and animals, in which a model is learned gradually using samples ordered in a meaningful sequence. A curriculum specifies a scheme under which training samples will be gradually learned. CL has successfully improved the performance on a variety of problems. In our problem, our intuition is that a curriculum, similar to its role in education, may provide meaningful supervision to help a student overcome corrupted labels. A reasonable curriculum can help the student focus on the samples whose labels have a high chance of being correct.

However, for the deep CNNs, we need to address two limitations of the existing CL methodology. First, existing curriculums are usually predefined and remain fixed during training, ignoring the feedback from the student. The learning procedure of deep CNNs is quite complicated, and may not be accurately modeled by the predefined curriculum. Second, the alternating minimization, commonly used in CL and self-paced learning~\cite{kumar2010self} requires alternative variable updates, which is difficult for training very deep CNNs via mini-batch stochastic gradient descent.

To this end, we propose a method to learn the curriculum from data by a network called \emph{MentorNet}. MentorNet learns a data-driven curriculum to supervise the base deep CNN, namely \emph{StudentNet}. MentorNet can be learned to approximate an existing predefined curriculum or discover new data-driven curriculums from data. The learned data-driven curriculum can be updated a few times taking into account of the StudentNet's feedback. Whenever MentorNet is learned or updated, we fix its parameter and use it together with StudentNet to minimize the learning objective, where MentorNet controls the timing and attention to learn each sample. At the test time, StudentNet makes predictions alone without MentorNet.

The proposed method improves existing curriculum learning in two aspects. First, our curriculum is learned from data rather than predefined by human experts. It takes into account of the feedback from StudentNet and can be dynamically adjusted during training. Intuitively, this resembles a ``collaborative'' learning paradigm, where the curriculum is determined by the teacher and student together. Second, in our algorithm, the learning objective is jointly minimized using MentorNet and StudentNet via mini-batch stochastic gradient descent. Therefore, the algorithm can be conveniently parallelized to train deep CNNs on big data. We show the convergence and empirically verify it on large-scale benchmarks.

We verify our method on four benchmarks. Results show that it can significantly improve the performance of deep CNNs trained on both controlled and real-world corrupted training data. Notably, to the best of our knowledge, it achieves the best-published result on WebVision~\cite{li2017webvision}, a large benchmark containing 2.2 million images of real-world noisy labels. To summarize, the contribution of this paper is threefold:

\begin{itemize}
\vspace{-3mm}
    \item We propose a novel method to learn data-driven curriculums for deep CNNs trained on corrupted labels.
\vspace{-5mm}
    \item We discuss an algorithm to perform curriculum learning for deep networks via mini-batch stochastic gradient descent.
\vspace{-2mm}
    \item We verify our method on 4 benchmarks and achieve the best-published result on the WebVision benchmark.
\vspace{-5mm}
\end{itemize}

\vspace{-5mm}
\section{Preliminary on Curriculum Learning}
We formulate our problem based on the model in~\cite{kumar2010self} and~\cite{jiang2015self}. Consider a classification problem with the training set $\mathcal{D} = \{(\mathbf{x}_1,\mathbf{y}_1),\cdots, (\mathbf{x}_n,\mathbf{y}_n)\}$, where $\mathbf{x}_i$ denotes the $i^{th}$ observed sample and $\mathbf{y}_i \in \{0,1\}^m$ is the noisy label vector over $m$ classes. Let $g_{s}(\mathbf{x}_i,\mathbf{w})$ denote the discriminative function of a neural network called \emph{StudentNet}, parameterized by $\mathbf{w} \in \mathbb{R}^d$. Further, let $\mathbf{L}(\mathbf{y}_i,g_s(\mathbf{x}_i,\mathbf{w}))$, a $m$-dimensional column vector, denote the loss over $m$ classes. Introduce the latent weight variable, $\mathbf{v} \in \mathbb{R}^{n \times m}$, and optimize the objective:
\begin{align}
\label{eq:datareg_obj}
\min_{\mathbf{w}  \in \mathbb{R}^d ,\mathbf{v}\in [0,\!1]^{n \times m}}\!\!\mathbb{F}(\mathbf{w},\!\mathbf{v})
&= \nonumber\\[-1.0ex]
\frac{1}{n}\sum_{i=1}^n \mathbf{v}_i^T \mathbf{L}(\mathbf{y}_i,\!g_s(\mathbf{x}_i,\!\mathbf{w})) + G(\mathbf{v}; \lambda) + \theta \|\mathbf{w}\|_2^2
\end{align}
where $\|\cdot\|_2$ is the $l_2$ norm for weight decay, and data augmentation and dropout are subsumed inside $g_s$. $\mathbf{v}_i \in [0,1]^{m \times 1}$ is a vector to represent the latent weight variable for the $i$-th sample. The function $G$ defines a \emph{curriculum}, parameterized by $\lambda$. This paper focuses on the one-hot label. For notation convenience, denote the loss $\mathbf{L}(\mathbf{y}_i,\!g_s(\mathbf{x}_i,\!\mathbf{w}))\!=\!\ell_i$, $\mathbf{v}_i$ as a scalar $v_i$, and $\mathbf{y}_i$ as an integer $y_i \in [1,m]$. 

In the existing literature, alternating minimization~\cite{csiszar1984information}, or its related variants, is commonly employed to minimize the training objective, e.g. in~\cite{kumar2010self,ma2017self,jiang2014self}. This is an algorithmic paradigm where $\mathbf{w}$ and $\mathbf{v}$ are alternatively minimized, one at a time while the other is held fixed. When $\mathbf{v}$ is fixed, the weighted loss is typically minimized by stochastic gradient descent. When $\mathbf{w}$ is fixed, we compute $\mathbf{v}^{k} = \arg\min_{\mathbf{v}} \mathbb{F} (\mathbf{v}^{k-1}, \mathbf{w}^{k})$ using the most recently updated $\mathbf{w}^{k}$ at epoch $k$. For example, Kumar \etal~\yrcite{kumar2010self} employed $G(\mathbf{v}) = - \lambda \|\mathbf{v}\|_1$. When $\mathbf{w}$ is fixed, the optimal $\mathbf{v}$ can be easily derived by:
\vspace{-2mm}
\begin{equation}
\label{eq:selfpaced_sol_v}
v_i^* = \mathds{1}(\ell_i \le \lambda), \forall i \in [1,n],
\vspace{-2mm}
\end{equation}
where $\mathds{1}$ is the indicator function. Eq.~\eqref{eq:selfpaced_sol_v} intuitively explains the predefined curriculum in~\cite{kumar2010self}, known as self-paced learning. First, when updating $\mathbf{v}$ with a fixed $\mathbf{w}$, a sample of smaller loss than the threshold $\lambda$ is treated as an ``easy" sample, and will be selected in training ($v_i^*=1$). Otherwise, it will not be selected ($v_i^*=0$). Second, when updating $\mathbf{w}$ with a fixed $\mathbf{v}$, the classifier is trained only on the selected ``easy" samples. The hyperparameter $\lambda$ controls the learning pace and corresponds to the ``age'' of the model. When $\lambda$ is small, only samples of small loss will be considered. As $\lambda$ grows, more samples of larger loss will be gradually added to train a more ``mature'' model.

As shown, the function $G$ specifies a curriculum, \ie, a sequence of samples with their corresponding weights to be used in training. When $\mathbf{w}$ is fixed, its optimal solution, \eg~ Eq.~\eqref{eq:selfpaced_sol_v}, computes the time-varying weight that controls the timing and attention to learn every sample. Recent studies discovered multiple predefined curriculums and verified them in many real-world applications, \eg, in~\cite{fan2017self,ma2017self,sangineto2016self,fan2017self,chang2017active}.

This paper studies learning curriculum from data. In the rest of this paper, Section~\ref{sec:mentornet} presents an approach to learn data-driven curriculum by \emph{MentorNet}. Section~\ref{sec:algorithm} discusses an algorithm to optimize Eq.~\eqref{eq:datareg_obj} using MentorNet and StudentNet together via mini-batch training.

\vspace{-3mm}
\section{Learning Curriculum from Data}\label{sec:mentornet}
Existing curriculums are either predetermined as an analytic expression of $G$ or a function to compute sample weights. Such predefined curriculums cannot be adjusted accordingly, taking into account of the feedback from the student. This section discusses a new way to learn data-driven curriculum by a neural network, called \emph{MentorNet}. The MentorNet $g_m$ is learned to compute time-varying weights for each training sample. Let $\Theta$ denote the parameters in $g_m$. Given a fixed $\mathbf{w}$, our goal is to learn an $\Theta^*$ to compute the weight:
\vspace{-1mm}
\begin{equation}
\label{eq:gm_obj}
g_m(\mathbf{z}_i; \Theta^*) = \arg\min_{v_i \in [0,1]} \mathbb{F} (\mathbf{w}, \mathbf{v}), \forall i \in [1,n]
\vspace{-2mm}
\end{equation}
where $\mathbf{z}_i = \phi(\mathbf{x}_i, y_i, \mathbf{w})$ indicates the input feature to MentorNet about the $i$-th sample.

\subsection{Learning Curriculum}\label{sec:learning_curriculum}
MentorNet can be learned to 1) approximate existing curriculums or 2) discover new curriculums from data. 

\textbf{Learning to approximate predefined curriculums.} Our first task is to learn a MentorNet to approximate a predefined curriculum. To do so, we minimize the objective in Eq.~\eqref{eq:datareg_obj}:
\vspace{-2mm}
\begin{equation}
\label{eq:mentornet_learning_obj_explicit}
\arg\min_{\Theta}  
\sum_{(\mathbf{x}_i,y_i) \in \mathcal{D}} g_m (\mathbf{z}_i;\Theta) \ell_i +  G(g_m (\mathbf{z}_i; \Theta); \lambda)
\vspace{-1mm}
\end{equation}
Eq.~\eqref{eq:mentornet_learning_obj_explicit} applies for both convex and non-convex $G$. This paper employs the following predefined curriculum. It is derived from~\cite{jiang2015self} and works well in our experiments. As will be discussed later, it is also related to robust non-convex penalties. 
\vspace{-3mm}
\begin{equation}
\label{eq:predefined_curriculum}
G(\mathbf{v}; \lambda) = \sum_{i=1}^n  \frac{1}{2} \lambda_2 v_i^2  - (\lambda_1+\lambda_2) v_i,
\vspace{-1mm}
\end{equation}
where $\lambda_1, \lambda_2 \ge 0$ are hyper-parameters. As $G$ is convex, there exists a closed-form solution for the optimal value of Eq.~\eqref{eq:gm_obj}. Given a fixed $\mathbf{w}$, define $\mathbb{F}_{\mathbf{w}}(\mathbf{v}) = \sum_{i=1}^n f(v_i)$:
\vspace{-2mm}
\begin{equation}
f(v_i) = v_i \ell_i + \frac{1}{2} \lambda_2 v_i^2 - (\lambda_1+\lambda_2) v_i
\vspace{-2mm}
\end{equation}
The minima are obtained at $\nabla_{\mathbf{v}} \mathbb{F}_{\mathbf{w}} (\mathbf{v}) = 0$, and can be decoupled by setting $\partial f / \partial v_i = 0$. We then have:
\begin{equation}
\label{eq:predefined_closed_form}
g_m(\mathbf{z}_i; \Theta^*) =\begin{cases}
\mathds{1}(\ell_i \le \lambda_1)  &  \lambda_2 = 0 \\
\min(\max (0, 1- \frac{\ell_i - \lambda_1}{\lambda_2}), 1) & \lambda_2 \ne 0\\
\end{cases},
\end{equation}
where $\Theta^*$ is the optimal MentorNet parameter obtained by SGD. The closed-form solution in Eq.~\eqref{eq:predefined_closed_form} gives some intuitions about the curriculum. When $\lambda_2=0$, it is similar to self-paced learning~\cite{kumar2010self} \ie~only ``easy'' samples of $\ell_i < \lambda_1$ will be selected in training ($g_m(\mathbf{z}_i; \Theta^*)=1$). When $\lambda_2 \ne 0$, samples of loss $\ell_i \ge \lambda_2 + \lambda_1$ will not be selected in training. These samples represent the ``hard'' samples of greater loss. Otherwise, samples will be weighted linearly w.r.t. $1-(\ell_i-\lambda_1)/\lambda_2$. As in~\cite{kumar2010self}, the hyper-parameters $\lambda_1$ and $\lambda_2$ control the learning pace.

\textbf{Learning data-driven curriculums.} Our next task is to learn a curriculum solely derived from labeled data. To this end, $\Theta$ is learned on another dataset $\mathcal{D}' = \{(\phi(\mathbf{x}_i, y_i, \mathbf{w}), v_i^*)\}$, where $(\mathbf{x}_i, y_i)$ is sampled from $\mathcal{D}$ and $|\mathcal{D}'| \ll |\mathcal{D}|$. $v_i^*$ is a given annotation and we assume it approximates the optimal weight, \ie, $v_i^* \simeq \arg\min_{v_i \in [0,1]} \mathbb{F} (\mathbf{v}, \mathbf{w})$. In this paper, we assign binary labels to $v_i^*$, where $v_i^*=1$ iff $y_i$ is a correct label. As $v_i^*$ is binary, $\Theta$ is learned by minimizing the cross-entropy loss between $v_i^*$ and $g(\mathbf{z}_i; \Theta)$. Intuitively, this process is similar to a mock test for the teacher (MentorNet) to learn to update her teaching strategy (curriculum). The student (StudentNet) provides features $\phi(\cdot, \cdot, \mathbf{w})$ for the mock test using the latest model $\mathbf{w}$. The teacher can learn an updated curriculum from the data to better supervise the latest student model. The learned curriculum is jointly determined by the teacher and student together.

The information on the correct label may not always be available on the target dataset $\mathcal{D}$. In this case, we learn the curriculum on a different small dataset where the correct labels are available. Intuitively, it resembles first learning a teaching strategy with the student on one topic and transfer the strategy on a similar topic. Empirically, Section~\ref{sec:exp_part2} substantiates that the learned curriculum on a small subset of CIFAR-10 can be applied to the target CIFAR-100 dataset.

A burn-in period is introduced before learning $\Theta$. In the first 20\% training epoch of the StudentNet, MentorNet is initialized and fixed as $g_m(\mathbf{z}_i ;\Theta^*)=r_i$, where $r_i \sim \text{Bernoulli}(p)$ is the Bernoulli random variable. This is equivalent to randomly dropping out $p$\% training samples. We found that the burn-in process helps StudentNet stabilize the prediction and focus on learning simple and common patterns.

\begin{figure}[ht]
\vspace{-3mm}
\centering
\includegraphics[width=0.9\linewidth]{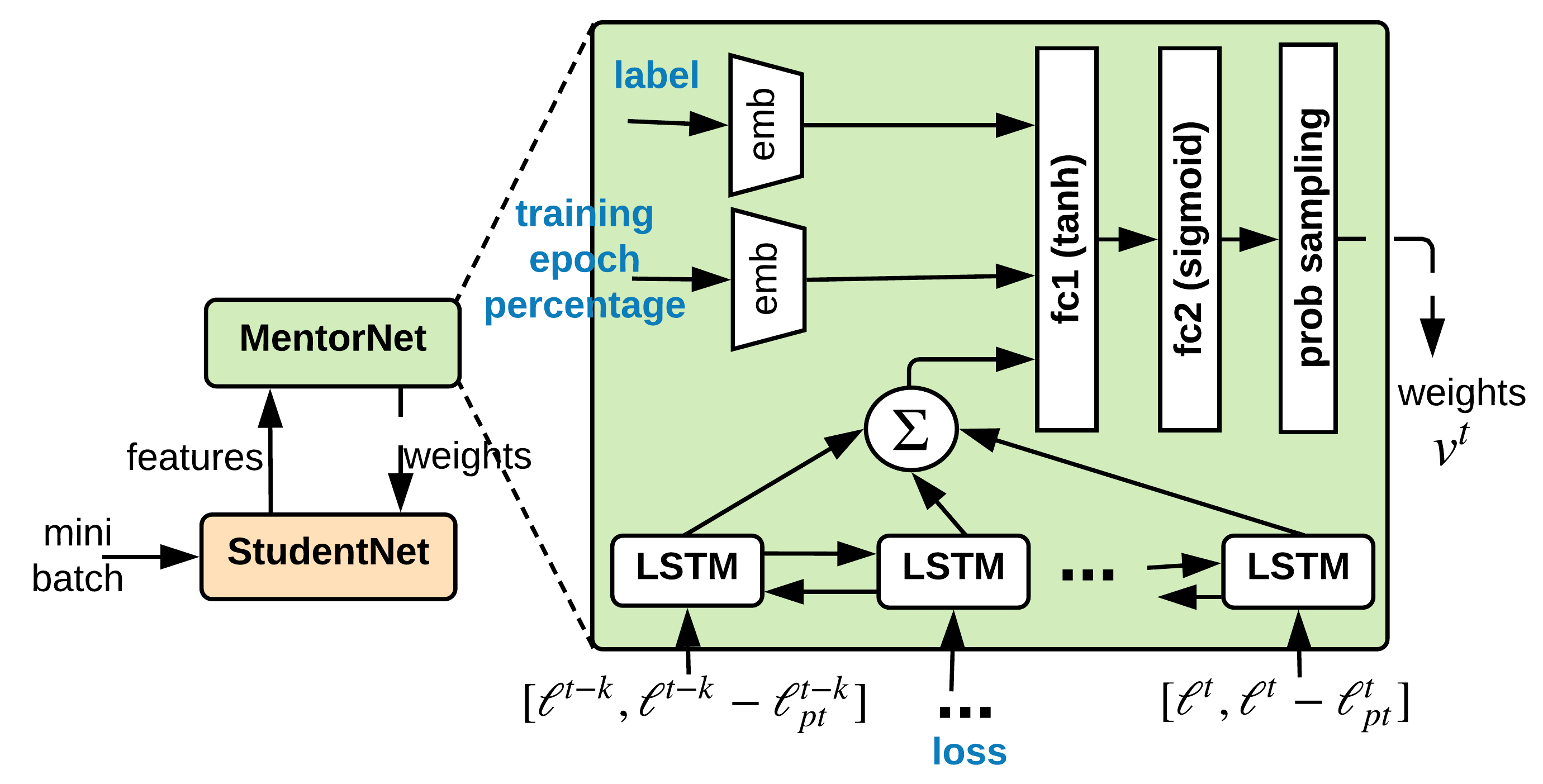}
\vspace{-5mm}
\caption{\label{fig:mentornet_arch}The MentorNet architecture used in experiments. \textit{emb}, \textit{fc} and \textit{prob sampling} stand for the embedding, fully-connected and probabilistic sampling layer.}
\vspace{-5mm}
\end{figure}

\textbf{MentorNet architecture.} We found that MentorNet can have a simple architecture. Appendix D shows that even MentorNet based on the two-layer perceptron can reasonably approximate the existing curriculum in the literature. Nevertheless, we use a MentorNet architecture shown in Fig.~\ref{fig:mentornet_arch}, which works reasonably well compared to classical network architectures. It takes the input of a mini-batch of samples, and outputs their corresponding sample weights. The feature $\mathbf{z}_i = \phi(\mathbf{x}_i, y_i, \mathbf{w})$ includes the loss, loss difference to the moving average, label and epoch percentage. $\ell_{pt}$ maintains an exponential moving average on the $p$-th percentile of the loss in each mini-batch. For a sample, its loss $\ell$ and loss difference $\ell-\ell_{pt}$ over the last few epochs can be encoded by a bidirectional LSTM network to capture the prediction variance~\cite{chang2017active}. We verify the LSTM encoder in the experiments in Appendix D. For simplicity, we set the step size of the LSTM to 1 in Section~\ref{sec:exp_part2} and only consider the loss and the loss difference of the current epoch.

The label and the training epoch percentage are encoded by two separate embedding layers. The epoch percentage is represented as an integer between 0 and 99. It is used to indicate the StudentNet's training progress, where 0 represents the first and 99 represents the last training epoch. The concatenated outputs from the LSTM and the embedding layers are fed into two fully-connected layers $fc_1, fc_2$, where $fc_2$ uses the sigmoid activation to ensure the output weights bounded between 0 and 1. The last layer in Fig.~\ref{fig:mentornet_arch} is a probabilistic sampling layer, and is used to implement the sample dropout in the burn-in process on the already learned MentorNet. 

\vspace{-2mm}
\subsection{Discussions}\label{sec:robust_learning}
MentorNet is a general framework for both predefined and data-driven curriculum learning, where various curriculums can be learned by the same MentorNet structure with different parameters. This framework is conceptually general and practically flexible as we can switch curriculums by attaching different MentorNets without modifying the pipeline. Therefore, we also learn MentorNets for predefined curriculums. For predefined curriculums where $G$ is unknown, we directly minimize the error between the MentorNet's outputs and desired weights. For example, the desired weight for focal loss~\cite{lin2017focal} is computed by: 
\begin{equation}
\label{eq:focal_loss_weight}
v_i^* = [1-\exp\{-\ell_i\}]^\gamma,
\vspace{-2mm}
\end{equation}
where $\gamma$ is a hyperparameter for smoothing the distribution.

This paper tackles the problem of overcoming corrupted labels. It is interesting to analyze why the learned curriculum can improve the generalization performance. It turns out that StudentNet, when jointly learned with MentorNet, may optimize an underlying robust objective and the objective is also related to the robust M-estimator~\cite{huber2011robust}.

To show this, let $v^*(\lambda,x)$ represent the optimal weight function for a loss variable $x$, and we define:
\begin{equation}
\label{eq:v_shorthand}
v^*(\lambda, x) = \text{argmin}_{v \in [0,1]} \ \ v x + G(v, \lambda).
\vspace{-1mm}
\end{equation}
As $g_m$ is an approximator to Eq.~\eqref{eq:v_shorthand}, its property can then be analyzed by the function $v^*(\lambda, x)$. Meng \etal \yrcite{meng2015objective} investigated the insights of self-paced objective function, and proved that the optimization of SPL algorithm is intrinsically equivalent to minimizing a robust loss function. They showed that given a fixed $\lambda$ and a decreasing $v^*(\lambda, x)$ with respect to $x$, the underlying objective of Eq.~\eqref{eq:datareg_obj} can be obtained by:
\begin{equation}
\label{eq:latent_f_lambda}
F_\lambda(\mathbf{w}) = \frac{1}{n} \sum_{i=1}^n \int_0^{\ell_i}v^*(\lambda,x) d x,
\vspace{-3mm}
\end{equation}

Based on it, the underlying learning objective of the curriculum in Eq.~\eqref{eq:predefined_curriculum} can then be derived.
\begin{remark}
When $\lambda_1, \lambda_2$ are fixed and $\lambda_2 \ne 0$, the underlying objective function of the curriculum in Eq.~\eqref{eq:predefined_curriculum} is calculated from:
\vspace{-1mm}
\begin{equation}
\label{eq:f_lambda_robust}
F_\lambda (\mathbf{w}) \!=\! \frac{1}{n}\sum_{i=1}^n\!\begin{cases}
\ell_i & \ell_i \le \lambda_1\\
(\lambda_2 + 2\lambda_1)/ 2 & \ell_i \ge \lambda_2 + \lambda_1 \\
\theta \ell_i \!-\! \ell_i^2 /(2 \lambda_2) \!-\! \frac{(\theta-1)^2 \lambda_2}{2} & \text{otherwise}\\
\end{cases}
\vspace{-1mm}
\end{equation}
where $\theta = (\lambda_2+\lambda_1)/\lambda_2$. When $\theta=1$ it is equivalent to the minimax concave penalty~\cite{zhang2010nearly}.
\vspace{-2mm}
\end{remark}

As shown in Eq.~\eqref{eq:f_lambda_robust}, the underlying objective has a form of $F_\lambda(\mathbf{w}) = \sum_i \rho(\ell_i)/n$, where $\rho$ is the penalty function in M-estimator~\cite{candes2008enhancing}. Particularly, when $\theta=1$, $\rho(\ell)$ is equivalent to the minimax concave plus penalty~\cite{zhang2010nearly}, a popular non-convex robust loss. The result indicates the learned MentorNet that approximates our predefined curriculum in Eq.~\eqref{eq:predefined_curriculum} leads to an underlying robust objective of the StudentNet.

For the data-driven curriculum, if the learned MentorNet satisfies certain conditions, we have:
\begin{proposition}
Suppose $(\mathbf{x}, y)$ denotes a training sample and its corrupted label. For simplicity, let the MentorNet input $\phi(\mathbf{x}, y, \mathbf{w}) = \ell$ be the loss computed by the StudentNet model parameter $\mathbf{w}$. The MentorNet $g_m(\ell;\Theta) = v$, where $v$ is the sample weight. If $g_m$ decreases with respect to $\ell$, then there exists an underlying robust objective $F$:
\vspace{-2mm}
\[
F(\mathbf{w}) = \frac{1}{n} \sum_{i=1}^n \rho(\ell_i),
\vspace{-3mm}
\]
where $\rho(\ell_i)  = \int_0^{\ell_i} g_m(x;\Theta) d x$. In the special cases, $\rho(\ell)$ degenerates to the robust M-estimator: Huber~\cite{huber1964robust} and the log-sum
penalty~\cite{candes2008enhancing}.
\end{proposition}


\begin{figure}[ht]
\vspace{-3mm}
\centering
\includegraphics[width=0.8\linewidth]{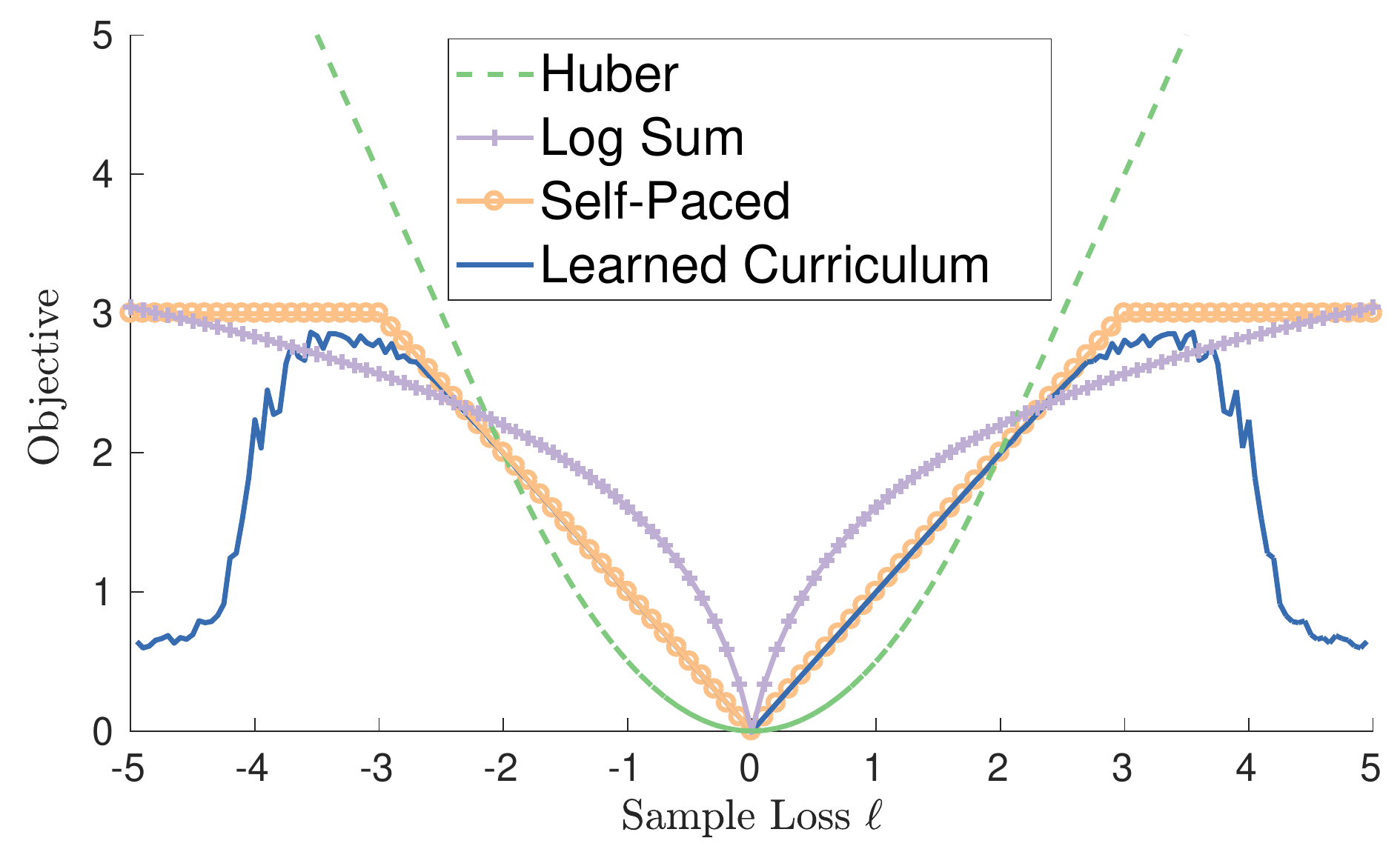}
\vspace{-3mm}
\caption{\label{fig:loss_curve}Visualization of sample loss and learning objective. The learned curriculum represents the best data-driven curriculum found in experiments.}
\vspace{-2mm}
\end{figure}

The proposition indicates that there exist some learned MentorNets that are related to the robust M-estimator. On noisy data, the effect of the robust objective is evident, \ie, preventing StudentNet from being dominated by corrupted labels. Fig.~\ref{fig:loss_curve} visualizes curves of the sample loss $\ell=y_i-g_s(\mathbf{x}_i, \mathbf{w})$ and the learning objective for the Huber loss~\cite{huber1964robust}, log-sum penalty~\cite{candes2008enhancing}, self-paced~\cite{kumar2010self}, and our learned data-driven curriculum. We use the best learned curriculum $\Theta^*$ on CIFAR-10 in our experiments and plot $|g_m(\phi(\mathbf{x},y,\mathbf{w});\Theta^*) \times \ell|$ since the $G$ in the objective function is unknown. As shown, all curves are robust to great loss to different extents. The corrupted labels in our problem are harmful. As the sample loss grows bigger beyond some value, MentorNet starts to sharply decrease the sample's weight. The subtlety of learned curriculum is difficult to be predefined by the analytic expression. Proposition 1 does not guarantee there is an underlying robust objective for every learned MentorNet. Instead, it shows MentorNet's capability of learning such robust objective. 

\vspace{-2mm}
\section{The Algorithm}\label{sec:algorithm}
\vspace{-1mm}

The alternating minimization algorithm~\cite{csiszar1984information} used in related work is intractable for deep CNNs, especially on big datasets, for two important reasons. First, in the subroutine of minimizing $\mathbf{w}$ when fixing $\mathbf{v}$, stochastic gradient descent often takes many steps before converging. This means that it can take a long time before moving past this single sub-step. However, such computation is often wasteful, particularly in the initial part of training, because, when $\mathbf{v}$ is far away from the optimal point, there is not much gain in finding the exact optimal $\mathbf{w}$ corresponding to this $\mathbf{v}$. Second, the subroutine of minimizing $\mathbf{v}$ when fixing $\mathbf{w}$ is often difficult, because the fixed vector $\mathbf{v}$ may not only consume a considerable amount of the memory but also hinder the parallel training on multiple machines. Therefore, optimizing the objective with deep CNNs requires some thought on the algorithmic level.

To minimize Eq.~\eqref{eq:datareg_obj}, we propose an algorithm called \emph{SPADE} (Scholastic gradient PArtial DEscent). The algorithm optimizes the StudentNet model parameter $\mathbf{w}$ jointly with a given MentorNet. It provides a simple and elegant way to minimize $\mathbf{w}$ and $\mathbf{v}$ stochastically over mini-batches. As a general approach, it can also take an input of $G$. Let $\Xi_t = \{(\mathbf{x}_j, y_j)\}_{j=1}^b$ denotes a mini-batch of $b$ samples, fetched uniformly at random and $\mathbf{v}_{\Xi}^{t} = [v_1^{t}, ... , v_b^{t}]$ represent the sample weights in $\Xi_t$. The MentorNet computes:
\vspace{-1mm}
\begin{equation}
\mathbf{v}_{\Xi}^{t} \!=\! g_m (\phi(\Xi_t, \mathbf{w}^{t-1})) \!=\! \arg\min_{\mathbf{v}_{\Xi}} \mathbb{F} (\mathbf{w}^{t-1},\! \mathbf{v}^{t-1}),
\vspace{-1mm}
\end{equation}
where $\phi$ is the feature extraction function defined in Eq.~\eqref{eq:gm_obj}. $\Theta$ denotes the learned MentorNet discussed in Section~\ref{sec:learning_curriculum}.

As shown in Algorithm~\ref{alg:overall}, for $\mathbf{w}$, a stochastic gradient is computed (via a mini-batch) and applied (Step 12), where $\alpha_t$ is the learning rate. For the latent weight variables $\mathbf{v}$, gradient descent is only applied to a small subset thereof parameters corresponding only to the mini-batch (Step 9 or 11). The partial gradient update on weight parameters is performed when $G$ is used (Step 9). Otherwise, we directly apply the weights computed by the learned MentorNet (Step 11). In both cases, the weights are computed on-the-fly within a mini-batch and thus do not need to be fixed. As a result, the algorithm can be conveniently parallelized across multiple machines.

{
\vspace{-2mm}
\begin{algorithm}[ht]
\footnotesize
\SetKwData{Left}{left}\SetKwData{This}{this}\SetKwData{Up}{up}
\SetKwInOut{Input}{Input}\SetKwInOut{Output}{Output}
\LinesNumbered
\Input{Dataset $\mathcal{D}$, a predefined $G$ or a learned $g_m(\cdot;\Theta)$}
\Output{The model parameter $\mathbf{w}$ of StudentNet.}
Initialize $\mathbf{w}^{0}, \mathbf{v}^{0}$, $t = 0$\;

\While{Not Converged}{
	Fetch a mini-batch $\Xi_t$ uniformly at random\;
	
	For every $(\mathbf{x}_i, y_i)$ in $\Xi_t$ compute $\phi(\mathbf{x}_i, y_i, \mathbf{w}^t)$\;
	
	\If{update curriculum}{
	    $\Theta \leftarrow \Theta^*$, where $\Theta^*$ is learned in Sec.~\ref{sec:learning_curriculum}
	}
	
    \If{$G$ is used}{
	    $\mathbf{v}_\Xi^{t} \leftarrow  \mathbf{v}_\Xi^{t-1} - \alpha_t  \nabla_{\mathbf{v}}
	    \mathbb{F}(\mathbf{w}^{t-1}, \mathbf{v}^{t-1})\rvert_{\Xi_t}$
	}\lElse{
	    $\mathbf{v}_\Xi^{t} \leftarrow g_m(\phi(\Xi_t, \mathbf{w}^{t-1}); \Theta)$
    }

	$\mathbf{w}^{t}  \leftarrow \mathbf{w}^{t-1} - \alpha_t  \nabla_{\mathbf{w}} \mathbb{F}(\mathbf{w}^{t-1}, \mathbf{v}^{t}) \rvert_{\Xi_t}$\;
	
    $t \leftarrow t + 1$
}
\Return $\mathbf{w}^{t}$
\caption{{\small SPADE for minimizing Eq.~\eqref{eq:datareg_obj}}}
\label{alg:overall}
\end{algorithm}
\vspace{-2mm}
}

The curriculum can change during training. MentorNet is updated a few times in Algorithm~\ref{alg:overall}. In Step 6, the MentorNet parameter $\Theta$ is updated to adapt to the most recent model parameters of StudentNet. In experiments, we update $\Theta$ twice after the learning rate is changed. Each time, a data-driven curriculum is learned from the data generated by the most recent $\mathbf{w}$ using the method discussed in Section~\ref{sec:learning_curriculum}. The update is consistent with existing curriculum learning methodology~\cite{bengio2009curriculum,kumar2010self} and the difference here is that for each update, the curriculum is learned rather than specified by human experts.

Under standard assumptions, Theorem 1 shows that the algorithm stabilizes and converges to a stationary point (convergence to global/local minima cannot be guaranteed unless in specially structured non-convex objectives~\cite{chen2018gradient,zhou2017stochastic,zhou2017mirror}). The proof is in Appendix B. The theorem is a characterization of stability of the model parameters $\mathbf{w}$. For the weight parameters $\mathbf{v}$, as it is restricted in a compact set, convergence to a stationary point is not always guaranteed. As the model parameters are more important, we only provide a detailed characterization of the model parameter.

\begin{theorem}
\label{theorem:1}
Let the objective $\mathbb{F}(\mathbf{w}, \mathbf{v})$ defined in Eq.~\eqref{eq:datareg_obj} be differentiable, $L(\cdot)$ be Lipschitz continuous in $\mathbf{w}$ and $\nabla_{\mathbf{v}}G(\cdot)$ be Lipschitz continuous in $\mathbf{v}$.
Let $\mathbf{w}^{t}, \mathbf{v}^{t}$ be iterates from Algorithm 1 and $\sum_{t=0}^{\infty} \alpha_t = \infty, \sum_{t=0}^{\infty} \alpha_t^2 < \infty$ . Then,
$\lim_{t\rightarrow \infty} \mathbb{E}[\|\nabla_{\mathbf{w}} \mathbb{F}(\mathbf{w}^{t}, \mathbf{v}^{t})\|_2^2] = 0$.
\vspace{-2mm}
\end{theorem}

For the manually designed curriculums, it may be unclear where or even whether such predefined curriculum would converge via mini-batch training. Theorem~\ref{theorem:1} shows that the learned curriculum can converge and produce a stable StudentNet model. The algorithm can be used to replace the alternating minimization method in related work.

\vspace{-3mm}
\section{Experiments}
This section empirically verifies the proposed method on four benchmarks of controlled corrupted labels in Section~\ref{sec:exp_part2} and real-world noisy labels in Section~\ref{sec:exp_part3}. The code can be found at {\small \url{https://github.com/google/mentornet}}.

\vspace{-2mm}
\subsection{Experiments on controlled corrupted labels}\label{sec:exp_part2}
\vspace{-2mm}
This section validates MentorNet on the controlled corrupted label. We follow a common setting in~\cite{zhang2017understanding} to train deep CNNs, where the label of each image is independently changed to a uniform random class with probability $p$, where $p$ is noise fraction and is set to 0.2, 0.4 and 0.8. The labels of validation data remain clean for evaluation.

\noindent\textbf{Dataset and StudentNet}: We use the same benchmarks in~\cite{zhang2017understanding}: CIFAR-10, CIFAR-100 and ImageNet. CIFAR-10 and CIFAR-100~\cite{krizhevsky2009learning} consist of 32 $\times$ 32 color images arranged in 10 and 100 classes. Both datasets contain 50,000 training and 10,000 validation images. ImageNet ILSVRC2012~\cite{deng2009imagenet} contain about 1.2 million training and 50k validation images, split into 1,000 classes. Each image is resized to 299x299 with 3 color channels.

We employ 3 recent deep CNNs as our StudentNets: inception~\cite{szegedy2016rethinking}, resnet-101~\cite{he2016deep} with wide filters~\cite{zagoruyko2016wide} and inception-resnet v2~\cite{szegedy2017inception}. Table~\ref{tab:studentnet} shows their \#model parameters, training, and validation accuracy when we train them on the clean training data (noise$=0$). As shown, they achieve reasonable accuracy on each task.

\begin{table}[ht]
\vspace{-5mm}
\centering
\footnotesize
\caption{\label{tab:studentnet}StudentNet and their accuracies on the clean training data.}
\begin{tabular}{cl|ccc}
\hline
\multicolumn{1}{l}{Dataset} & Model             & \#para & {\footnotesize train acc} & {\footnotesize val acc} \\
\hline
\multirow{2}{*}{CIFAR10}    & {\footnotesize inception}     &1.7M         & 0.83         & 0.81        \\
                            & {\footnotesize resnet101}         & 84M       & 1.00         & 0.96        \\
\hline
\multirow{2}{*}{CIFAR100}   & {\footnotesize inception}     &1.7M              &0.64                &0.49               \\
                            & {\footnotesize resnet101}        & 84M   & 1.00         & 0.79        \\
\hline
\multirow{1}{*}{ImageNet}   & {\footnotesize inception\_resnet}  &59M             & 0.88         & 0.77        \\
\hline
\end{tabular}
\vspace{-3mm}
\end{table}


\begin{table*}[ht]
\vspace{-4mm}
\centering
\caption{Comparison of validation accuracy on CIFAR-10 and CIFAR-100 under different noise fractions.}
\label{tab:cifar_comparison}
\begin{tabular}{|l|ccc|ccc|ccc|ccc|}
\hline
                & \multicolumn{6}{c|}{Resnet-101 StudentNet}  & \multicolumn{6}{c|}{Inception StudentNet}\\
                & \multicolumn{3}{c|}{CIFAR-100} & \multicolumn{3}{c|}{CIFAR-10} & \multicolumn{3}{c|}{CIFAR-100} & \multicolumn{3}{c|}{CIFAR-10} \\
Method & 0.2      & 0.4     & 0.8     & 0.2      & 0.4      & 0.8     & 0.2      & 0.4     & 0.8     & 0.2      & 0.4      & 0.8     \\
\hline
FullModel&0.60 & 0.45 & 0.08 &0.82 & 0.69 & 0.18 & 0.43 & 0.38 & 0.15 & 0.76 & 0.73 & 0.42\\
Forgetting&0.61&0.44&0.16&0.78&0.63&0.35&0.42&0.37&0.17&0.76&0.71&0.44\\
Self-paced& 0.70 & 0.55  &0.13 & 0.89&0.85&0.28 &0.44&0.38&0.14&\textbf{0.80}&0.74&0.33 \\     
Focal Loss&0.59&0.44&0.09&0.79&0.65&0.28&0.43&0.38&0.15&0.77&0.74&0.40 \\
Reed Soft&0.62&0.46&0.08&0.81&0.63&0.18&0.42&0.39&0.12&0.78&0.73&0.39\\
MentorNet PD &0.72&0.56&0.14&0.91&0.77&0.33&0.44&0.39&0.16&0.79&0.74&0.44\\
MentorNet DD &\textbf{0.73}& \textbf{0.68}  & \textbf{0.35} &  \textbf{0.92} & \textbf{0.89} &\textbf{0.49} &\textbf{0.46}& \textbf{0.41}&\textbf{0.20}&0.79&\textbf{0.76}&\textbf{0.46}\\
\hline
\end{tabular}
\vspace{-4mm}
\end{table*}

\noindent \textbf{Baselines}: MentorNet is compared against the following baselines: \textbf{FullMode} is the standard StudentNet trained using $l_2$ weight decay, dropout~\cite{srivastava2014dropout} and data augmentation~\cite{krizhevsky2012imagenet}. The hyper-parameters are set to the best ones found on the clean training data. Unless specified otherwise, for a fair comparison, the StudentNet with the same hyperparameters is used in all baseline and our model. \textbf{Forgetting} was introduced in~\cite{arpit2017closer}, in which the dropout parameter is searched in the range of (0.2-0.9). \textbf{Self-paced}~\cite{kumar2010self} and \textbf{Focal Loss}~\cite{lin2017focal} represent well-known predefined curriculums in the literature. We implemented \textbf{Reed}~\yrcite{reed2014training} and \textbf{Goldberger}~\cite{goldberger2017training} as the recent weakly-supervised learning methods. The above baseline methods are a mixture of the curriculum learning and the recent methods dealing with corrupted labels.
 
\noindent \textbf{Our Model}: \textbf{MentorNet PD} is the network learned using our \emph{predefined} curriculum in Eq.~\eqref{eq:predefined_curriculum} using no additional clean labels. \textbf{MentorNet DD} is the learned \emph{data-driven} curriculum. It is trained on 5,000 images of true labels, randomly sampled from the CIFAR-10 training set. The same data are used to learn MentorNet DD on CIFAR-100. Note CIFAR-10 and CIFAR-100 are two different datasets that have not only different classes but also the different number of classes. Therefore, it is fair to compare MentorNet DD with other methods using no true labels on CIFAR-100. Algorithm~\ref{alg:overall} is used to optimize the StudentNet. The decay factor in computing the loss moving average is set to 0.95. The loss percentile in the moving average is set by the cross-validation. As mentioned, a burn-in process is used in the first 20\% training epoch for both MentorNet DD and MentorNet PD. More details are discussed in Appendix E.

We first show the comparison to the baseline method on CIFAR-10 and CIFAR-100 in Table~\ref{tab:cifar_comparison}. On both datasets, each method is verified with two StudentNets (resnet-101 and inception) under the noise fraction of 0.2, 0.4, and 0.8. As we see on both datasets, MentorNet improves FullModel across different noise fractions, and the learned data-driven curriculum (MentorNet DD) achieves the best results. The improvement is more significant for the deeper CNN model resnet-101. For example, on the CIFAR-10 of 40\% noise, MentorNet DD (with resnet-101) yields an absolute 20\% gain over FullModel. After inspecting the result, we found that it may be because Mentor DD learns a more appropriate curriculum to give high weights to samples of correct labels. As a result, it helps the StudentNet focus on samples of correct labels. The results indicate that the learned MentorNet can improve the generalization performance of recent deep CNNs, and outperform the predefined curriculums (Self-paced and Focal Loss).

\begin{figure}[ht]
\vspace{-2mm}
\centering
\subfigure[{Test error on clean validation}]
{
\label{fig:exp_generation_a}
\includegraphics[width=0.7\linewidth, height=28mm]{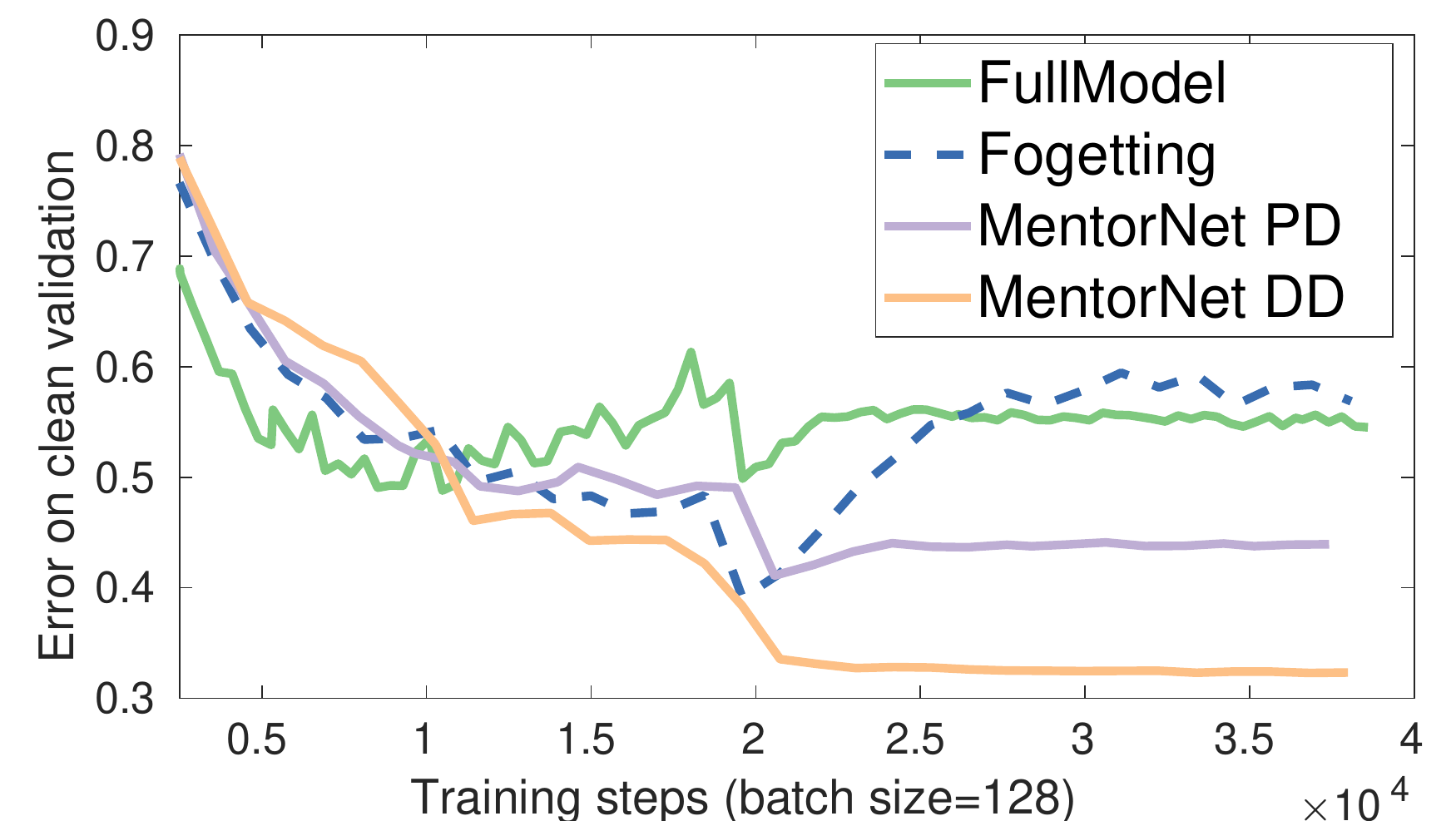}
\vspace{-5mm}
}
\subfigure[ {Training error on corrupted labels}]
{
\label{fig:exp_generation_b}
\includegraphics[width=0.7\linewidth, height=28mm]{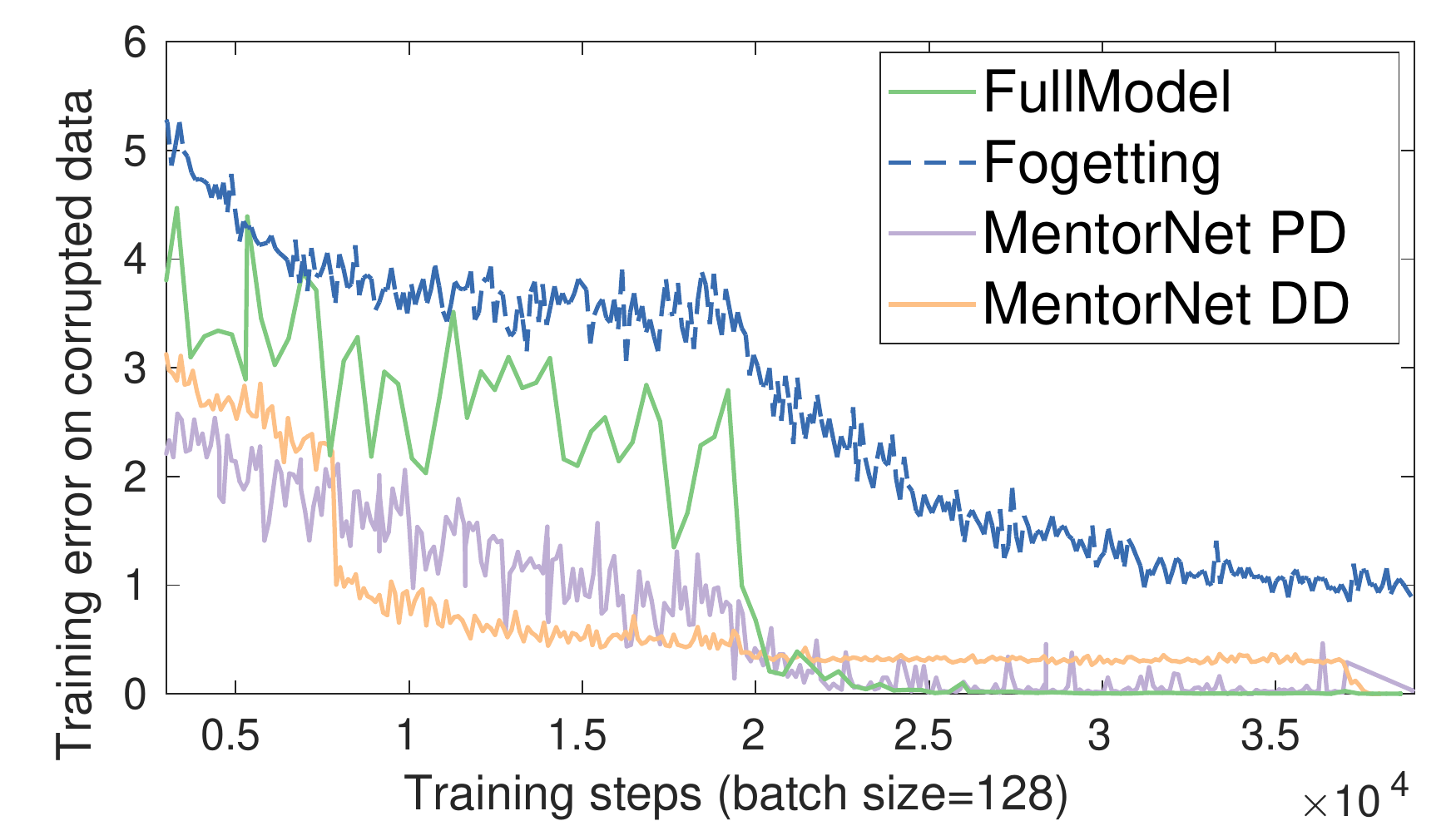}
\vspace{-5mm}
}
\vspace{-5mm}
\caption{\label{fig:exp_generation}Error of resnet trained on CIFAR-100 of 40\% noise.}
\vspace{-7mm}
\end{figure}

Fig.~\ref{fig:exp_generation} plots the training and test error on the clean validation data, under a representative setting: resnet-101 on CIFAR-100 of 40\% noise, where the $x$-axis denotes the training iteration. The $y$-axis is the validation error on the clean validation in Fig.~\ref{fig:exp_generation_a} and the mini-batch training error on corrupted labels in Fig.~\ref{fig:exp_generation_b}. For MentorNet, the training error is computed by $\sum_i v_i \ell_i$. The figure shows two insights. First, the training error of MentorNet approaches zero. This empirically verifies the convergence of the model. Second, MentorNet can overcome the overfitting to the corrupted label. While the training error is decreasing, the test error does not increase in Fig.~\ref{fig:exp_generation_a}. It suggests that the learned curriculum is beneficial for StudentNet. The sharp change around the 20k iteration in Fig.~\ref{fig:exp_generation} is due to the learning rate change. Besides, our result is consistent with~\cite{zhang2017understanding} that deep CNNS is able to get 0 training error on the corrupted training data. Forgetting (the dashed curve) is the only one that does not converge within 30k steps. As indicated in~\cite{arpit2017closer}, it is because forgetting reduces the speed at which DNNs memorize. As suggested in~\cite{zhang2017mixup}, a \emph{not} converged model might yield a better result, \eg, stop the model at 20K in Fig.~\ref{fig:exp_generation}. However, as it is hard to predetermine the time for early stopping, our focus is comparing the converged model. 

\begin{figure}[ht]
\vspace{-1mm}
\includegraphics[width=1\linewidth]{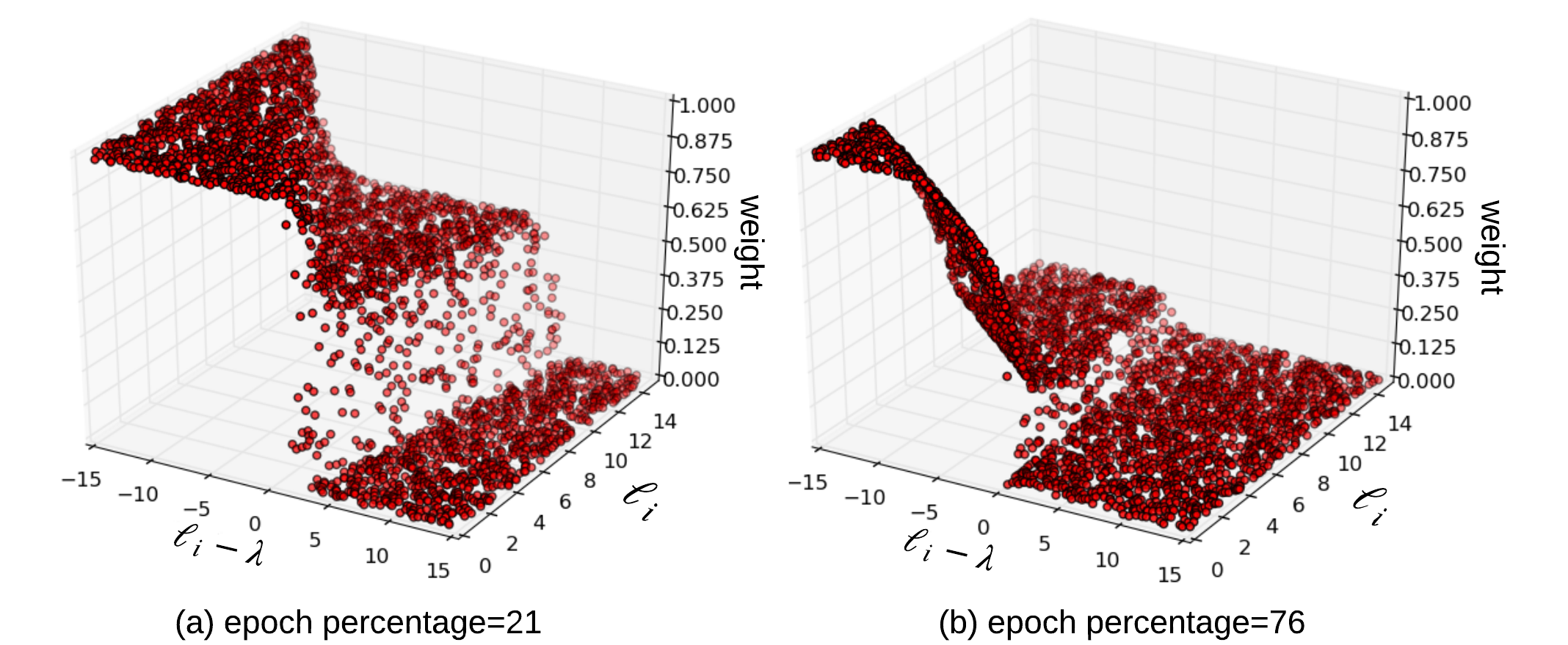}
\vspace{-6mm}
\caption{\label{fig:visual_curriculum} The data-driven curriculums learned by MentorNet with the resnet-101 at epoch 21 in (a) and 76 in (b).}
\vspace{-5mm}
\end{figure}

Fig.~\ref{fig:visual_curriculum} illustrates the best learned data-driven curriculum in our experiments, where the $z$-axis denotes the weights computed by $g_m$; the $y$ and $x$ axes denote the sample loss and the loss difference to the moving average, where $\lambda$ is the loss moving average. Two observations can be found in Fig.~\ref{fig:visual_curriculum}. First, the learned curriculum changes during the training of the StudentNet. Fig.~\ref{fig:visual_curriculum} (a) and (b) are MentorNet learned at different epochs. As shown, (a) assigns greater weights to samples of big loss more aggressively. Second, the learned curriculums in Fig.~\ref{fig:visual_curriculum} generally satisfy the condition in Proposition 1, \ie, the weight generally decreases with the loss. It suggests that joint learning of StudentNet and MentorNet optimizes an underlying robust objective.

Table~\ref{tab:exp_compare_soa} compares to recent published results under the setting: CIFAR of 40\% noise fraction. We cite the number in~\cite{azadi2016auxiliary}, and implement other methods using the same resnet-101 StudentNet. The results show that our result is comparable and even better than the state-of-the-art.

\begin{table}[ht]
\vspace{-5mm}
\centering
\small
\caption{\label{tab:exp_compare_soa} Validation accuracy comparison to representative published results on CIFAR of 40\% noise.}
\vspace{1mm}
\begin{tabular}{ll|cc}
\hline
ID&Method & CIFAR-10 & CIFAR-100 \\
\hline
1&Reed Hard (Resnet)  &  0.62       & 0.47  \\
2&Reed Soft (Resnet)&  0.61       & 0.46  \\
3&Goldberger (Resnet) &  0.70       & 0.46  \\
4&Azadi (AlexNet)~\yrcite{azadi2016auxiliary} &0.75& -\\
5&\textbf{MentorNet (Resnet)} &\textbf{0.89}&\textbf{0.68}\\
\hline
\end{tabular}
\vspace{-3mm}
\end{table}

To verify MentorNet for large-scale training, we apply our method on the ImageNet ILSVRC12~\cite{deng2009imagenet} benchmark to improve the inception-resnet v2~\cite{szegedy2017inception} model. We train the model on the ImageNet of 40\% noise. Inspired by~\cite{zhang2017understanding}, we start with an inception-resnet (NoReg) with no regularization (NoReg) and add weight decay, dropout, and data augmentation to the model. Table~\ref{tab:exp_imagenet} shows the comparison. As shown, MentorNet improves the performance of both the inception-resnet without regularization (NoReg) and with full regularization (FullModel). It also outperforms the forgetting baseline (dropout keep probability = 0.2). The results suggest that MentorNet can improve deep CNNs on the large-scale training on corrupted labels.

\begin{table}[ht]
\vspace{-3mm}
\centering
\caption{\label{tab:exp_imagenet} Accuracy on the clean ImageNet ImageNet validation set. Models are trained on the ImageNet training data of 40\% noise.}
\small
\begin{tabular}{l|cc}
\hline
Method & P@1 & P@5 \\
\hline
NoReg &  0.538 & 0.770  \\
NoReg+WeDecay &  0.560 & 0.809  \\
NoReg+Dropout & 0.575&0.807\\
NoReg+DataAug&  0.522 & 0.772  \\
NoReg+MentorNet&0.590& 0.814\\
\hline
FullModel&  0.612 & 0.844  \\
Forgetting(FullModel) &0.628&0.845\\
\textbf{MentorNet(FullModel)} &\textbf{0.651}&\textbf{0.859}\\
\hline
\end{tabular}
\vspace{-5mm}
\end{table}

\vspace{-2mm}
\subsection{Experiments on real-world noisy labels}
\label{sec:exp_part3}
\vspace{-2mm}
To verify MentorNet on real-world noisy labels, we conduct experiments on the large WebVision benchmark~\cite{li2017webvision}. It contains 2.4 million images of real-world noisy labels, crawled from the web using the 1,000 concepts in ImageNet ILSVRC12. We download the resized images from the official website\footnote{https://www.vision.ee.ethz.ch/webvision/download.html}. The inception-resenet v2~\cite{szegedy2017inception} is used as our StudentNet, trained using a distributed asynchronized momentum optimizer on 50 GPUs. Since the dataset is very big, for quick experiments, we compare baseline methods using the Google image subset on the first 50 classes. We use Mini to denote this subset and Entire for the entire WebVision. All the models are evaluated on the clean ILSVRC12 and WebVision validation set.

Table~\ref{tab:exp_real_world_noise} lists the comparison result. As we see, the proposed MentorNet significantly improves baseline methods on real-world noisy labels. The method marked by the start indicates it uses a pre-trained ImageNet model to obtain additional 30k labels for 118 classes. Following the same protocol, MentorNet* is trained using the additional labels. The results show that our method outperforms the baseline methods on real-world noisy labels. To the best of our knowledge, it achieves the best-published result on the WebVision~\cite{li2017webvision} benchmark.

\begin{table}[ht]
\vspace{-5mm}
\caption{\label{tab:exp_real_world_noise}Validation accuracy on the ImageNet ILSVRC12 and WebVision validation set. The number outside (inside) the parentheses denotes top-1 (top-5) classification accuracy (\%). * marks the method trained using additional verification labels.}
\centering
\small
\begin{tabular}{cl|cc}
\hline
Dataset& Method & ILSVRC12 & WebVision\\
\hline
Entire& Li \etal~\yrcite{li2017webvision} &0.476 (0.704)&0.570 (0.779)\\
Entire& Forgetting &0.590 (0.808) & 0.666 (0.856)\\
Entire& Lee \etal~\yrcite{lee2017cleannet}* &0.602 (0.811) & 0.685 (0.865)\\
Entire& MentorNet &0.625 (0.830) & 0.708 (0.880)\\
Entire& \textbf{MentorNet*} &\textbf{0.642} (\textbf{0.848})& \textbf{0.726} (\textbf{0.889})\\
\hline
Mini&FullModel &0.585 (0.818)&-\\
Mini&Forgetting &0.562 (0.816)&-\\
Mini&Reed Soft&0.565 (0.827)&-\\
Mini&Self-paced &0.576 (0.822)& -\\
Mini& \textbf{MentorNet} &\textbf{0.638} (\textbf{0.858})&-\\
\hline
\end{tabular}
\vspace{-5mm}
\end{table}

\vspace{-1mm}
\section{Related Work}
\vspace{-1mm}
Curriculum learning (CL), proposed by Bengio \etal~\yrcite{bengio2009curriculum}, is a learning paradigm in which a model is learned by gradually including from easy to complex samples in training so as to increase the learning entropy~\cite{bengio2009curriculum}. From the human behavioral perspective, Khan \etal~\yrcite{khan2011humans} have shown that CL is consistent with the principle of human teaching. CL has been empirically verified in a variety of problems, such as computer vision~\cite{supancic2013self,chen2015webly}, natural language processing~\cite{turian2010word}, multitask learning~\cite{graves2017automated}. A common CL approach is to predefine a curriculum. For example, Kumar \etal~\yrcite{kumar2010self} proposed a curriculum called self-paced learning which favors training samples of smaller loss. After that, many predefined curriculums were proposed, \eg, in \cite{supancic2013self,jiang2014self,jiang2015self,sangineto2016self,chang2017active,ma2017self,ma2017convergence}. For example, Jiang \etal~\yrcite{jiang2014self} introduced a curriculum of using easy and diverse samples. Fan \etal~\yrcite{fan2017self} proposed to use predefined sample weighting schemes as an implicit way to define a curriculum. Previous work has shown that predefined curriculums are useful in overcoming noisy labels~\cite{chen2015webly,liang2016learning,lin2017active}. In parallel to CL, the sample weighting schemes were also studied in~\cite{lin2017active,wang2017robust,fan2018learning,dehghani2018fidelityweighted}. Compared to the existing work, our paper presents a new way of learning data-driven curriculums for deep networks trained on corrupted labels.


Our work is related to the weakly-supervised learning methods. Among recent contributions, Reed \etal~\yrcite{reed2014training} developed a robust loss to model ``prediction consistency''. Menon \etal~\yrcite{menon2015learning} used class-probability estimation to study the corruption process. Sukhbaatar \etal~\yrcite{sukhbaatar2014training} proposed a noise transformation to estimate the noise distribution. The transformation matrix needs to be periodically updated and is non-trivial to learn. To address the issue, Goldberger \etal~\yrcite{goldberger2017training} proposed to add an additional softmax layer end-to-end with the base model. Azadi \etal~\yrcite{azadi2016auxiliary} tackled this problem by a regularizer called AIR. This method was shown to be effective but it relied on additional clean labels to train the representation.
More recently, methods utilized additional labels for label cleaning~\cite{Veit2017}, knowledge distillation~\cite{Li2017ICCV} or semi-supervised learning~\cite{vahdat2017toward,dehghani2017avoiding}. Different from previous work, we focus on learning curriculum to train very deep CNNs on corrupted labels from scratch. In addition, clean labels are not always needed for our method. In Section~\ref{sec:exp_part2}, the MentorNet is learned on a small subset of CIFAR-10 and applied to CIFAR-100

\vspace{-2mm}
\section{Conclusions}
\vspace{-2mm}
In this paper, we presented a novel method for training deep CNNs on corrupted labels. Our work was built on curriculum learning and advanced the methodology by proposing to learn data-driven curriculum via a neural network called MentorNet. We proposed an algorithm for jointly optimizing deep CNNs with MentorNet on large-scale data. We conducted comprehensive experiments on datasets of controlled and real-world noise. Our empirical results showed that generalization performance of deep CNNs trained on corrupted labels can be effectively improved by the learned data-driven curriculum.

\section*{Acknowledgements}
The authors would like to thank anonymous reviewers for helpful comments and Deyu Meng, Sergey Ioffe, and Chong Wang for meaningful discussions and kind support.

\bibliography{example_paper}
\bibliographystyle{icml2018}

\balance
\end{document}


\onecolumn

\icmltitle{Supplementary Materials:\\
MentorNet Learning Data-Driven Curriculum \\
for Very Deep Neural Networks on Corrupted Labels}

\begin{icmlauthorlist}
\icmlauthor{Lu Jiang}{}
\icmlauthor{Zhengyuan Zhou}{}
\icmlauthor{Thomas Leung}{}
\icmlauthor{Li-Jia Li}{}
\icmlauthor{Li Fei-Fei}{}
\end{icmlauthorlist}

\icmlcorrespondingauthor{Lu Jiang}{lujiang@google.com}


\appendix

\section{Derivation of Remark 1}
Our objective function is:
\begin{equation}
\begin{split}
\label{eq:datareg_obj}
\min_{\mathbf{w}  \in \mathbb{R}^d ,\mathbf{v}\in [0,\!1]^{n}}\!\!\mathbb{F}(\mathbf{w},\!\mathbf{v})
= \frac{1}{n}\sum_{i=1}^n v_i \mathbf{L}(y_i,\!g_s(\mathbf{x}_i,\!\mathbf{w})) + G(\mathbf{v}; \lambda) + \theta \|\mathbf{w}\|_2^2
\vspace{-5mm}
\end{split}
\end{equation}

Let $\ell_i = \mathbf{L}(y_i,g_s(\mathbf{x}_i,\mathbf{w}))$ denote the loss of the $i$-th sample ($\ell_i \ge 0$). The predefined curriculum is defined as:
\begin{equation}
G(\mathbf{v}; \lambda) = \sum_{i=1}^n  \frac{1}{2} \lambda_2 v_i^2  - (\lambda_2+\lambda_1) v_i.
\end{equation}

Denote $\mathbb{F}_{\mathbf{w}}$ as the objective function when the $\mathbf{w}$ is fixed. We have
\begin{equation}
\begin{split}
\mathbb{F}_{\mathbf{w}}(\mathbf{v}) &= \frac{1}{n} \sum_{i=1}^n v_i \ell_i  +   G(\mathbf{v}; \lambda) + \theta \|\mathbf{w}\|_2^2 \\
&= \frac{1}{n} \sum_{i=1}^n v_i \ell_i + \frac{1}{2} \lambda_2 v_i^2  - (\lambda_2+\lambda_1) v_i + \theta \|\mathbf{w}\|_2^2 \\
&= \frac{1}{n} \sum_{i=1}^n f(v_i) + \theta \|\mathbf{w}\|_2^2
\end{split}
\end{equation}
where $f(v_i) = v_i \ell_i + \frac{1}{2} \lambda_2 v_i^2 - (\lambda_2+\lambda_1) v_i$.

As $f(v_i)$ is convex with respect to $v_i$ ($\lambda_2 \ge 0$). Its minimum is obtained at
\begin{equation}
\begin{split}
&\arg\min_{\mathbf{v} \in [0,1]^n} \nabla_{\mathbf{v}} \mathbb{F}_{\mathbf{w}} (\mathbf{v}) = 0  \\
&\Rightarrow \frac{\partial f(v_i)}{\partial v_i} = \ell_i + \lambda_2 v_i - \lambda_2 - \lambda_1 = 0, \text{ } (\forall i \in [1,n], v_i \in [0,1])
\end{split}
\end{equation}
Since $\lambda_1, \lambda_2 \ge 0$ and $v_i$ is bounded in $[0,1]$. When $\lambda_2 \ne 0$, the optimal $v_i^*$ is calculated from:
\begin{equation}
\label{eq:vstar1}
v_i^* =\begin{cases}
1 &  (\ell_i \le \lambda_1) \land (\lambda_2 \ne 0)\\
1- \frac{ \ell_i - \lambda_1}{\lambda_2} &  (\lambda_1 < \ell_i < \lambda_2 + \lambda_1) \land (\lambda_2 \ne 0)\\
0 & (\ell_i \ge \lambda_2 + \lambda_1) \land (\lambda_2 \ne 0)
\end{cases},
\end{equation}
When $\lambda_2 = 0$, the optimal weight writes as:
\begin{equation}
\label{eq:vstar2}
v_i^* =\begin{cases}
1 & (\ell_i <  \lambda_1) \land (\lambda_2 = 0)\\
0 & (\ell_i \ge \lambda_1) \land (\lambda_2 = 0)
\end{cases}.
\end{equation}

Combined Eq.~\eqref{eq:vstar1} and Eq.~\eqref{eq:vstar2}, we have
\begin{equation}
\begin{split}
\label{eq:vstar3}
v_i^*&=\begin{cases}
\mathds{1}(\ell_i \le \lambda_1)  &  \lambda_2 = 0 \\
\min(\max (0, 1- \frac{\ell_i-\lambda_1}{\lambda_2}),1) & \lambda_2 \ne 0 
\end{cases}
\end{split}
\end{equation}

According to the definition of $\Theta$,  we have $g_m(\phi(\mathbf{x}_i, y_i, \mathbf{w}); \Theta^*)  = v_i^*$. Incorporating Eq.~\eqref{eq:vstar3}, we have
\begin{equation}
\label{eq:gm_sol}
g_m(\phi(\mathbf{x}_i, y_i, \mathbf{w}); \Theta^*) =\begin{cases}
\mathds{1}(\ell_i \le \lambda_1)  &  \lambda_2 = 0 \\
\min(\max (0, 1- \frac{\ell_i-\lambda_1}{\lambda_2}), 1) & \text{otherwise}
\end{cases}
\end{equation}

Now we derive its underlying objective. First, we define a function $v^*(\lambda, x)$ and incorporate the above optimal solution:
\begin{equation}
v^*(\lambda, x) = \arg\min_{v \in [0,1]} \ \ v x + G(v, \lambda)
=\begin{cases}
\mathds{1}(x \le \lambda_1)  &  \lambda_2 = 0 \\
\min(\max (0, 1- \frac{x-\lambda_1}{\lambda_2}),1) & \text{otherwise}
\end{cases}
\end{equation}

Let $\epsilon > 0$ denote a small positive constant. Using the condition $\lambda_2 \ge 0$, we incorporate Eq.~\eqref{eq:vstar3}:
\begin{equation}
v^*(\lambda, x+\epsilon) - v^*(\lambda, x)  \le 0
\end{equation}

That indicates $v^*(\lambda, x)$ decreases with respect to $x$. According to~\cite{meng2015objective}, given the fixed hyperparameter $\lambda$ (\textit{i.e.}~$\lambda_1,\lambda_2$), its underlying objective function has the form of:
\begin{equation}
\label{eq:latent_f_meng}
F_\lambda(\mathbf{w}) = \frac{1}{n} \sum_{i=1}^n \int_0^{\ell_i}v^*(\lambda;x) d x,
\end{equation}

After incorporating Eq.~\eqref{eq:vstar1} and Eq.~\eqref{eq:vstar2} into Eq.~\eqref{eq:latent_f_meng}, we have when $\lambda_2=0$
\begin{equation}
F_\lambda(\mathbf{w}) = \frac{1}{n}\sum_{i=1}^n \min(\ell_i, \lambda_1)
\end{equation}

When $\lambda_2 \ne 0$, we have:
\begin{equation}
\label{eq:flambda_1}
\begin{split}
F_\lambda(\mathbf{w}) = \frac{1}{n}\sum_{i=1}^n \begin{cases}
\ell_i & \ell_i \le \lambda_1 \\
\ell_i+ \frac{\lambda_1}{\lambda_2} \ell_i - \frac{1}{2\lambda_2} \ell_i^2 - \frac{\lambda_1^2}{2 \lambda_2}& \lambda_1 < \ell_i < \lambda_2 + \lambda_1\\
\frac{\lambda_2+2\lambda_1}{2} & \ell_i \ge \lambda_2 + \lambda_1
\end{cases}
\\ = \frac{1}{n}\sum_{i=1}^n \begin{cases}
\ell_i & \ell_i \le \lambda_1\\
\theta \ell_i - \ell_i^2 /(2 \lambda_2) - \frac{(\theta-1)^2 \lambda_2}{2} & \lambda_1 < \ell_i < \lambda_2 + \lambda_1\\
(\lambda_2 + 2\lambda_1)/ 2 & \ell_i \ge \lambda_2 + \lambda_1 \\
\end{cases}
\end{split},
\end{equation}
where $\theta = (\lambda_2+\lambda_1)/\lambda_2$ and $\lambda_1, \lambda_2 \ge 0$. We $\lambda_2 \ne 0$, we have the Eq.(10) in the Remark 1 in the paper.
\begin{equation}
\begin{split}
F_\lambda(\mathbf{w}) = \frac{1}{n}\sum_{i=1}^n \begin{cases}
\ell_i & \ell_i \le \lambda_1\\
(\lambda_2 + 2\lambda_1)/ 2 & \ell_i \ge \lambda_2 + \lambda_1 \\
\theta \ell_i - \ell_i^2 /(2 \lambda_2) - \frac{(\theta-1)^2 \lambda_2}{2} & \text{otherwise}\\
\end{cases}
\end{split},
\end{equation}

When $\theta =1$ we have $\lambda_1=0$ and the above equation becomes:
\begin{equation}
\begin{split}
F_\lambda(\mathbf{w}) = \frac{1}{n}\sum_{i=1}^n \begin{cases}
\ell_i - \ell_i^2 /(2 \lambda_2) & \ell_i < \lambda_2\\
\lambda_2 / 2 & \ell_i \ge \lambda_2 \\
\end{cases}
\end{split}.
\end{equation}

The above equation is equivalent to the minimax concave penalty (MCP)~\cite{gong2013general,zhang2010nearly}. Below is its formula presented in~\cite{gong2013general} (with the regularization hyperparameter setting to 1):
\begin{equation}
\begin{split}
\int_0^{\ell} [1- \frac{x}{t}] d x  = \begin{cases}
\ell - \ell^2 /(2 t) & \ell < t\\
t / 2 & \ell \ge t
\end{cases}
\end{split},
\end{equation}


\section{Proof of Theorem 1}

\begin{theorem}
	\label{theorem:1}
	Let the objective $\mathbb{F}(\mathbf{w}, \mathbf{v})$ defined in Eq.~\eqref{eq:datareg_obj} be differentiable, $L(\cdot)$ be Lipschitz continuous in $\mathbf{w}$ and $\nabla_{\mathbf{v}}G(\cdot)$ be Lipschitz continuous in $\mathbf{v}$.
	Let $\mathbf{w}^{t}, \mathbf{v}^{t}$ be iterates from Algorithm 1 and $\sum_{t=0}^{\infty} \alpha_t = \infty, \sum_{t=0}^{\infty} \alpha_t^2 < \infty$ . Then,
	$\lim_{t\rightarrow \infty} \mathbb{E}[\|\nabla_{\mathbf{w}} \mathbb{F}(\mathbf{w}^{t}, \mathbf{v}^{t})\|_2^2] = 0$.
\end{theorem}


\begin{proof}
	First, by the definition of the objective function $\mathbb{F}(\mathbf{w}, \mathbf{v})$, it can be easily checked that $L(\cdot)$ being Lipschitz continuous in $\mathbf{w}$ implies that $\nabla_{\mathbf{w}}\mathbb{F}(\mathbf{w}, \mathbf{v})$ is a Lipschitz function in $\mathbf{w}$ for every $\mathbf{v}$. Similarly, 
	$\nabla_{\mathbf{v}}G(\cdot)$ being Lipschitz continuous in $\mathbf{v}$
	implies that $\nabla_{\mathbf{v}}\mathbb{F}(\mathbf{w}, \mathbf{v})$ is a Lipschitz function in $\mathbf{v}$ for all $\mathbf{w}$.
	Further, without loss of generality, we assume all the Lipschitz constants are (upper bounded by) $L$.
	
	Throughout the proof,  as in the main text, $n$ is the size of the training dataset. 
	Define the $n$-dimensional vector $e^k$ as
	$e^t_i = 1$ if $(x_i, y_i) \in \Xi_t$ and $0$ otherwise and denote by $\otimes$ the point-wise product operation between two vectors: $[a_1, a_2] \otimes [b_1, b_2] = [a_1b_1, a_2b_2]$. 
	When $G$ is used, the update in each iteration is two consecutive gradient steps as follows: 
\begin{eqnarray*}
	\mathbf{w}^{t+1}& = &\mathbf{w}^{t} - \alpha_t  \nabla_{\mathbf{w}} \mathbb{F}(\mathbf{w}^t, \mathbf{v}^{t}) \rvert_{\Xi_t}, \\
	\mathbf{v}^{t+1}& = &\mathbf{v}^{t} - \alpha_t  \nabla_{\mathbf{v}} \mathbb{F}(\mathbf{w}^{t+1}, \mathbf{v}^{t})\rvert_{\Xi_t}.\\
\end{eqnarray*}
Since the mini-batch $\Xi_t$ is draw uniformly at random, we can rewrite the update as:
\begin{eqnarray*}
	\mathbf{w}^{t+1}& = &\mathbf{w}^{t} - \alpha_t  [\nabla_{\mathbf{w}} \mathbb{F}(\mathbf{w}^k, \mathbf{v}^{t}) + \xi^t], \\
	\mathbf{v}^{t+1}& = &\mathbf{v}^{t} - \alpha_t  [e^t \otimes \nabla_{\mathbf{v}} \mathbb{F}(\mathbf{w}^{t+1}, \mathbf{v}^{t})],\\
\end{eqnarray*}
where $\xi^t = \nabla_{\mathbf{w}} \mathbb{F}(\mathbf{w}^t, \mathbf{v}^{t}) \rvert_{\Xi_t} - \nabla_{\mathbf{w}} \mathbb{F}(\mathbf{w}^k, \mathbf{v}^{t})$.
Note that both $\xi^t$ and $e^t$ are \textbf{iid} random variables with finite variance,
since $\Xi_t$ are drawn \textbf{iid} with a finite number ($b$) of samples.
Further, $\mathbb{E}[\nabla_{\mathbf{w}} \mathbb{F}(\mathbf{w}^t, \mathbf{v}^{t}) \rvert_{\Xi_t} - \nabla_{\mathbf{w}} \mathbb{F}(\mathbf{w}^k, \mathbf{v}^{t})] = 0$,
since samples are drawn uniformly at random.

By Lipschitz continuity of $\nabla_{\mathbf{w}}F(\mathbf{w}, \mathbf{v})$ (and $L$ being the Lipschitz constant), we obtain the following:
\begin{align*}
&\mathbb{F}(\mathbf{w}^{t+1}, \mathbf{v}^t) - \mathbb{F}(\mathbf{w}^{t}, \mathbf{v}^t) \le
\langle \nabla_\mathbf{w}\mathbb{F}(\mathbf{w}^{t}, \mathbf{v}^t), \mathbf{w}^{t+1} - \mathbf{w}^t\rangle + \frac{L}{2} \|\mathbf{w}^{t+1} - \mathbf{w}^{t}\|_2^2  \\
& = \langle \nabla_\mathbf{w}\mathbb{F}(\mathbf{w}^{t}, \mathbf{v}^t), -\alpha_t  [\nabla_{\mathbf{w}} \mathbb{F}(\mathbf{w}^t, \mathbf{v}^{t}) + \xi^t]\rangle
+ \frac{L}{2}\|\mathbf{w}^{t+1} - \mathbf{w}^{t}\|_2^2 \\
&  = -\alpha_t \{\|\nabla_\mathbf{w}\mathbb{F}(\mathbf{w}^{t}, \mathbf{v}^t)\|_2^2 +
\langle \nabla_\mathbf{w}\mathbb{F}(\mathbf{w}^{t}, \mathbf{v}^t), \xi^t\rangle\}
+ \frac{L}{2} \|\mathbf{w}^{t+1} - \mathbf{w}^{t}\|_2^2\\
&  = -\alpha_t \{\|\nabla_\mathbf{w}\mathbb{F}(\mathbf{w}^{t}, \mathbf{v}^t)\|_2^2 +
\langle \nabla_\mathbf{w}\mathbb{F}(\mathbf{w}^{t}, \mathbf{v}^t), \xi^t\rangle\}
+ \frac{L\alpha_t^2}{2} \|\nabla_{\mathbf{w}} \mathbb{F}(\mathbf{w}^t, \mathbf{v}^{t}) + \xi^t\|_2^2 \\
& = -(\alpha_t - \frac{L\alpha_t^2}{2}) \|\nabla_\mathbf{w}\mathbb{F}(\mathbf{w}^{t}, \mathbf{v}^t)\|_2^2 +
\frac{L\alpha_t^2}{2}\|\xi^t\|_2^2 - (\alpha_t - L\alpha_t^2) \langle \nabla_\mathbf{w}\mathbb{F}(\mathbf{w}^{t}, \mathbf{v}^t), \xi^t\rangle.\\
\end{align*}

Similarly, by the Lipschitz continuity of $\nabla_{\mathbf{v}}F(\mathbf{w}, \mathbf{v})$, we have:
\begin{align*}
&\mathbb{F}(\mathbf{w}^{t+1}, \mathbf{v}^{t+1}) - \mathbb{F}(\mathbf{w}^{t+1}, \mathbf{v}^t) \le
\langle \nabla_\mathbf{v}\mathbb{F}(\mathbf{w}^{t+1}, \mathbf{v}^t), \mathbf{v}^{t+1} - \mathbf{v}^t\rangle + \frac{L}{2} \|\mathbf{v}^{t+1} - \mathbf{v}^{t}\|_2^2  \\
& = \langle \nabla_\mathbf{v}\mathbb{F}(\mathbf{w}^{t+1}, \mathbf{v}^t), -\alpha_t e^t \otimes \nabla_{\mathbf{v}} \mathbb{F}(\mathbf{w}^{t+1}, \mathbf{v}^{t})\rangle
+ \frac{L}{2} \|\mathbf{v}^{t+1} - \mathbf{v}^{t}\|_2^2 \\
& = -\alpha_t \langle \nabla_\mathbf{v}\mathbb{F}(\mathbf{w}^{t+1}, \mathbf{v}^t),  e^k \otimes \nabla_{\mathbf{v}} \mathbb{F}(\mathbf{w}^{t+1}, \mathbf{v}^{t})\rangle
+ \frac{L\alpha_t^2}{2} \| e^t \otimes \nabla_{\mathbf{v}} \mathbb{F}(\mathbf{w}^{t+1}, \mathbf{v}^{t})  \|_2^2. \\
\end{align*}

Note that when $G$ is not used, the bound for $\mathbb{F}(\mathbf{w}^{t+1}, \mathbf{v}^t) - \mathbb{F}(\mathbf{w}^{t}, \mathbf{v}^t) $ is still the same, but the bound for $\mathbb{F}(\mathbf{w}^{t+1}, \mathbf{v}^{t+1}) - \mathbb{F}(\mathbf{w}^{t+1}, \mathbf{v}^t)$ is now simply $\mathbb{F}(\mathbf{w}^{t+1}, \mathbf{v}^{t+1}) - \mathbb{F}(\mathbf{w}^{t+1}, \mathbf{v}^t) \le 0$.

Combining the above two equations, we then have:
\begin{enumerate}
	\item If $G$ is used, \begin{align*}
	&\mathbb{F}(\mathbf{w}^{t+1}, \mathbf{v}^{t+1}) - \mathbb{F}(\mathbf{w}^{t}, \mathbf{v}^t) = 
	\mathbb{F}(\mathbf{w}^{t+1}, \mathbf{v}^{t+1}) - \mathbb{F}(\mathbf{w}^{t+1}, \mathbf{v}^t) + \mathbb{F}(\mathbf{w}^{t+1}, \mathbf{v}^t) - \mathbb{F}(\mathbf{w}^{t}, \mathbf{v}^t) \\
	&\le -(\alpha_t - \frac{L\alpha_t^2}{2}) \|\nabla_\mathbf{w}\mathbb{F}(\mathbf{w}^{t}, \mathbf{v}^t)\|_2^2 +
	\frac{L\alpha_t^2}{2}\|\xi^t\|_2^2 - (\alpha_t - L\alpha_t^2) \langle \nabla_\mathbf{w}\mathbb{F}(\mathbf{w}^{t}, \mathbf{v}^t), \xi^t\rangle\\
	&-\alpha_t \langle \nabla_\mathbf{v}\mathbb{F}(\mathbf{w}^{t+1}, \mathbf{v}^k),  e^t \otimes \nabla_{\mathbf{v}} \mathbb{F}(\mathbf{w}^{t+1}, \mathbf{v}^{t})\rangle
	+ \frac{L\alpha_t^2}{2} \| e^t \otimes \nabla_{\mathbf{v}} \mathbb{F}(\mathbf{w}^{t+1}, \mathbf{v}^{t})  \|_2^2. \\
	\end{align*}
	If $G$ is not used, \begin{align*}
	&\mathbb{F}(\mathbf{w}^{t+1}, \mathbf{v}^{t+1}) - \mathbb{F}(\mathbf{w}^{t}, \mathbf{v}^t) \le
 \mathbb{F}(\mathbf{w}^{t+1}, \mathbf{v}^t) - \mathbb{F}(\mathbf{w}^{t}, \mathbf{v}^t) \\
	&\le -(\alpha_t - \frac{L\alpha_t^2}{2}) \|\nabla_\mathbf{w}\mathbb{F}(\mathbf{w}^{t}, \mathbf{v}^t)\|_2^2 +
	\frac{L\alpha_t^2}{2}\|\xi^t\|_2^2 - (\alpha_t - L\alpha_t^2) \langle \nabla_\mathbf{w}\mathbb{F}(\mathbf{w}^{t}, \mathbf{v}^t), \xi^t\rangle\\
	\end{align*}
	
\end{enumerate}

Taking expectation of both sides and since $\mathbf{E}[\xi^t] = 0$, we have if $G$ is used:
\begin{align*}
&\mathbf{E}[\mathbb{F}(\mathbf{w}^{t+1}, \mathbf{v}^{t+1})] - \mathbf{E}[\mathbb{F}(\mathbf{w}^{t}, \mathbf{v}^t)]\\
& \le  -(\alpha_t - \frac{L\alpha_t^2}{2}) \mathbf{E}[\|\nabla_\mathbf{w}\mathbb{F}(\mathbf{w}^{t}, \mathbf{v}^t)\|_2^2] +
\frac{L\alpha_t^2}{2}\mathbf{E}[\|\xi^t\|_2^2] 
-\alpha_t \mathbf{E}[\langle \nabla_\mathbf{v}\mathbb{F}(\mathbf{w}^{t+1}, \mathbf{v}^t),  e^t \otimes \nabla_{\mathbf{v}} \mathbb{F}(\mathbf{w}^{t+1}, \mathbf{v}^{t})\rangle] \\
&+ \frac{L\alpha_t^2}{2} \mathbf{E}[\| e^t \otimes \nabla_{\mathbf{v}} \mathbb{F}(\mathbf{w}^{t+1}, \mathbf{v}^{t})  \|_2^2] \\
&=-(\alpha_t - \frac{L\alpha_t^2}{2}) \mathbf{E}[\|\nabla_\mathbf{w}\mathbb{F}(\mathbf{w}^{t}, \mathbf{v}^t)\|_2^2] +
  \frac{L\alpha_t^2}{2}\mathbf{E}[\|\xi^t\|_2^2] 
  -\frac{\alpha_t b}{n} \mathbf{E}[\|\nabla_\mathbf{v}\mathbb{F}(\mathbf{w}^{t+1}, \mathbf{v}^t)\|_2^2]\\
  &+ \frac{L\alpha_t^2}{2} \sum_{i=1}^n \mathbf{E}[\| e^t_i \frac{\partial_{v_i} \mathbb{F}(\mathbf{w}^{t+1}, \mathbf{v}^{t})}{\partial v_i}  \|_2^2] \\
  & \le 
  -(\alpha_t - \frac{L\alpha_t^2}{2}) \mathbf{E}[\|\nabla_\mathbf{w}\mathbb{F}(\mathbf{w}^{t}, \mathbf{v}^t)\|_2^2] +
  \frac{L\alpha_t^2}{2}\mathbf{E}[\|\xi^t\|_2^2] 
  -\frac{\alpha_t b}{n} \mathbf{E}[\|\nabla_\mathbf{v}\mathbb{F}(\mathbf{w}^{t+1}, \mathbf{v}^t)\|_2^2]\\
  &+ \frac{L\alpha_t^2}{2} \sum_{i=1}^n \mathbf{E}[\|\frac{\partial_{v_i} \mathbb{F}(\mathbf{w}^{t+1}, \mathbf{v}^{t})}{\partial v_i}  \|_2^2] \\
  & =  -\alpha_t \mathbf{E}[\|\nabla_\mathbf{w}\mathbb{F}(\mathbf{w}^{t}, \mathbf{v}^t)\|_2^2]  -\frac{\alpha_t b}{n} \mathbf{E}[\|\nabla_\mathbf{v}\mathbb{F}(\mathbf{w}^{t+1}, \mathbf{v}^t)\|_2^2]
  \\
  & +\frac{L\alpha_t^2}{2} \{\mathbf{E}[\|\nabla_\mathbf{w}\mathbb{F}(\mathbf{w}^{t}, \mathbf{v}^t)\|_2^2] +  \mathbf{E}[\|\nabla_\mathbf{v}\mathbb{F}(\mathbf{w}^{t+1}, \mathbf{v}^t)\|_2^2] +   \mathbf{E}[\|\xi^t\|_2^2].  \}
\end{align*}
 where the second equality follows from 
$\mathbf{E}[\langle \nabla_\mathbf{v}\mathbb{F}(\mathbf{w}^{t+1}, \mathbf{v}^t),  e^t \otimes \nabla_{\mathbf{v}} \mathbb{F}(\mathbf{w}^{t+1}, \mathbf{v}^{t})\rangle] = \frac{b}{n} \mathbf{E}[\|\nabla_\mathbf{v}\mathbb{F}(\mathbf{w}^{t+1}, \mathbf{v}^t)\|_2^2]$, which can be checked by a straightforward combinatorial argument. 

Following a similar chain of steps, we have the following bound if $G$ is not used:
$\mathbf{E}[\mathbb{F}(\mathbf{w}^{t+1}, \mathbf{v}^{t+1})] - \mathbf{E}[\mathbb{F}(\mathbf{w}^{t}, \mathbf{v}^t)] \le  -\alpha_t \mathbf{E}[\|\nabla_\mathbf{w}\mathbb{F}(\mathbf{w}^{t}, \mathbf{v}^t)\|_2^2]  +\frac{L\alpha_t^2}{2} \{\mathbf{E}[\|\nabla_\mathbf{w}\mathbb{F}(\mathbf{w}^{t}, \mathbf{v}^t)\|_2^2]  +   \mathbf{E}[\|\xi^t\|_2^2].  \}$
Since there are only a finite number of training samples, all the random quantities have bounded support and all the second moments are upper bounded, leading to $\mathbf{E}[\|\xi^t\|_2^2] < \infty, \mathbf{E}[\|\nabla_\mathbf{w}\mathbb{F}(\mathbf{w}^{t}, \mathbf{v}^t)\|_2^2] < \infty, \mathbf{E}[\|\nabla_\mathbf{v}\mathbb{F}(\mathbf{w}^{t+1}, \mathbf{v}^t)\|_2^2] < \infty$, and let $B$ be an upper bound on $\mathbf{E}[\|\nabla_\mathbf{w}\mathbb{F}(\mathbf{w}^{t}, \mathbf{v}^t)\|_2^2] +  \mathbf{E}[\|\nabla_\mathbf{v}\mathbb{F}(\mathbf{w}^{t+1}, \mathbf{v}^t)\|_2^2] +   \mathbf{E}[\|\xi^t\|_2^2]$. 

By telescoping, if $G$ is used, we have:
$\mathbf{E}[\mathbb{F}(\mathbf{w}^{T+1}, \mathbf{v}^{T+1})] - \mathbb{F}(\mathbf{w}^{0}, \mathbf{v}^0) = 
\sum_{k=0}^T \big\{ \mathbf{E}[\mathbb{F}(\mathbf{w}^{k+1}, \mathbf{v}^{k+1})] - \mathbf{E}[\mathbb{F}(\mathbf{w}^{k}, \mathbf{v}^k)]\big\} \\ \le  -\sum_{k=0}^T \alpha_k \mathbf{E}[\|\nabla_\mathbf{w}\mathbb{F}(\mathbf{w}^{k}, \mathbf{v}^k)\|_2^2]  -\sum_{k=0}^T\frac{\alpha_k b}{n} \mathbf{E}[\|\nabla_\mathbf{v}\mathbb{F}(\mathbf{w}^{k+1}, \mathbf{v}^k)\|_2^2] + \frac{LB}{2}\sum_{k=0}^T \alpha_k^2.$
Taking the limit $T \rightarrow \infty$ of both sides, we obtain:
\begin{align*}
& -\sum_{k=0}^{\infty} \alpha_k \mathbf{E}[\|\nabla_\mathbf{w}\mathbb{F}(\mathbf{w}^{k}, \mathbf{v}^k)\|_2^2]  -\frac{ b}{n}\sum_{k=0}^{\infty}\alpha_k\mathbf{E}[\|\nabla_\mathbf{v}\mathbb{F}(\mathbf{w}^{k+1}, \mathbf{v}^k)\|_2^2] + \frac{LB}{2}\sum_{k=0}^{\infty} \alpha_k^2 \\
& \ge \lim_{T \rightarrow \infty} \mathbf{E}[\mathbb{F}(\mathbf{w}^{T+1}, \mathbf{v}^{T+1})] - \mathbf{E}[\mathbb{F}(\mathbf{w}^{0}, \mathbf{v}^0)] 
\ge \min_{\mathbf{w}, \mathbf{v}} \mathbb{F}(\mathbf{w}, \mathbf{v})] - \mathbb{F}(\mathbf{w}^{0}, \mathbf{v}^0) > -\infty. \\
\end{align*}
Since $\sum_{k=0}^{\infty} \alpha_k^2 < \infty$,
the above inequality immediately implies that 
$\sum_{k=0}^{\infty} \alpha_k \mathbf{E}[\|\nabla_\mathbf{w}\mathbb{F}(\mathbf{w}^{k}, \mathbf{v}^k)\|_2^2]  < \infty$ and $\sum_{k=0}^{\infty}\alpha_k \mathbf{E}[\|\nabla_\mathbf{v}\mathbb{F}(\mathbf{w}^{k+1}, \mathbf{v}^k)\|_2^2]<\infty$.

If $G$ is not used, then by a similar argument, we have $\sum_{k=0}^{\infty} \alpha_k \mathbf{E}[\|\nabla_\mathbf{w}\mathbb{F}(\mathbf{w}^{k}, \mathbf{v}^k)\|_2^2]  < \infty$.

By Lemma A.5 in~\cite{mairal2013stochastic}, to show $\lim_{k\rightarrow \infty} \mathbf{E}[\|\nabla_{\mathbf{w}} \mathbb{F}(\mathbf{w}^k, \mathbf{v}^k)\|_2^2] = 0$,
since $\alpha_k$ is not summable,
it suffices to show $\bigg|\mathbf{E}[\|\nabla_\mathbf{w}\mathbb{F}(\mathbf{w}^{k+1}, \mathbf{v}^{k+1})\|_2^2] -
\mathbf{E}[\|\nabla_\mathbf{w}\mathbb{F}(\mathbf{w}^{k}, \mathbf{v}^k)\|_2^2]\bigg|
\le C \alpha_k$ for some constant $C$.
To do so, we first recall a useful fact: for any two vectors $\mathbf{a}, \mathbf{b}$ and any finite-dimensional vector norm $\|\cdot\|$, 
\begin{equation}\label{eq:help}
\Big|(\|a\| + \|b\|)(\|a\|-\|b\|)\Big|\le \|a+b\|\|a-b\|.
\end{equation}

We have:
\begin{align*}
& \bigg|\mathbf{E}[\|\nabla_\mathbf{w}\mathbb{F}(\mathbf{w}^{t+1}, \mathbf{v}^{t+1})\|_2^2] -
\mathbf{E}[\|\nabla_\mathbf{w}\mathbb{F}(\mathbf{w}^{t}, \mathbf{v}^t)\|_2^2]\bigg| \\
& = \Bigg| 
\mathbf{E} \bigg[
\big(\|\nabla_\mathbf{w}\mathbb{F}(\mathbf{w}^{t+1}, \mathbf{v}^{t+1})\|_2 + \|\nabla_\mathbf{w}\mathbb{F}(\mathbf{w}^{t}, \mathbf{v}^{t})\|_2 \big)   
\big(\|\nabla_\mathbf{w}\mathbb{F}(\mathbf{w}^{t+1}, \mathbf{v}^{t+1})\|_2 - \|\nabla_\mathbf{w}\mathbb{F}(\mathbf{w}^{t}, \mathbf{v}^{t})\|_2 \big) 
\bigg] \Bigg| \\
& \le
\mathbf{E} \bigg[
\Big|\|\nabla_\mathbf{w}\mathbb{F}(\mathbf{w}^{t+1}, \mathbf{v}^{t+1})\|_2 + \|\nabla_\mathbf{w}\mathbb{F}(\mathbf{w}^{t}, \mathbf{v}^{t})\|_2 \Big|  \cdot 
\Big|\|\nabla_\mathbf{w}\mathbb{F}(\mathbf{w}^{t+1}, \mathbf{v}^{t+1})\|_2 - \|\nabla_\mathbf{w}\mathbb{F}(\mathbf{w}^{t}, \mathbf{v}^{t})\|_2  \Big|\bigg] \\
& \le 
\mathbf{E} \bigg[
\Big\|\nabla_\mathbf{w}\mathbb{F}(\mathbf{w}^{t+1}, \mathbf{v}^{t+1}) + \|\nabla_\mathbf{w}\mathbb{F}(\mathbf{w}^{t}, \mathbf{v}^{t})\Big\|_2   \cdot 
\Big\|\nabla_\mathbf{w}\mathbb{F}(\mathbf{w}^{t+1}, \mathbf{v}^{t+1})\|_2 - \|\nabla_\mathbf{w}\mathbb{F}(\mathbf{w}^{t}, \mathbf{v}^{t})\|_2 \Big\|
\bigg]\\
& \le
2LB \mathbf{E}\Big[\|(\mathbf{w}^{t+1}, \mathbf{v}^{t+1})_- (\mathbf{w}^{t}, \mathbf{v}^{t})\|_2\Big] \\
& \le
2LB\alpha_t \mathbf{E}\Bigg[\Big\|\Big(\nabla_{\mathbf{w}}\mathbb{F}(\mathbf{w}^{t}, \mathbf{v}^{t})+ \xi^k, e^t \otimes \nabla_{\mathbf{w}}\mathbb{F}(\mathbf{w}^{t+1}, \mathbf{v}^{t})\Big)\Big\|_2\Bigg]\\
&= 
2LB\alpha_t \mathbf{E}\Bigg[\sqrt{\Big\|\nabla_{\mathbf{w}}\mathbb{F}(\mathbf{w}^{t}, \mathbf{v}^{t})+ \xi^k\Big\|_2^2} + \sqrt{\Big\| e^t \otimes \nabla_{\mathbf{w}}\mathbb{F}(\mathbf{w}^{t+1}, \mathbf{v}^{t}) \Big\|_2^2}\Bigg] \\
&\le
2LB\alpha_t 
\sqrt{
	\mathbf{E}\Bigg[\Big\|\nabla_{\mathbf{w}}\mathbb{F}(\mathbf{w}^{t}, \mathbf{v}^{t})+ \xi^k\Big\|_2^2\Bigg] + \mathbf{E}\Bigg[\Big\| e^t \otimes \nabla_{\mathbf{w}}\mathbb{F}(\mathbf{w}^{t+1}, \mathbf{v}^{t}) \Big\|_2^2\Bigg]} \\
&\le
2LB\alpha_k
\sqrt{
	2\mathbf{E}\Bigg[\Big\|\nabla_{\mathbf{w}}\mathbb{F}(\mathbf{w}^{t}, \mathbf{v}^{t})\Big\|_2^2\Bigg]+ 2\mathbf{E}\Bigg[\Big\|\xi^t\Big\|_2^2\Bigg] + \mathbf{E}\Bigg[\Big\| \nabla_{\mathbf{w}}\mathbb{F}(\mathbf{w}^{t+1}, \mathbf{v}^{t}) \Big\|_2^2\Bigg]} \\
&\le
2LB\alpha_t \sqrt{5B^2} = 2\sqrt{5}B^2L \alpha_t,
\end{align*}
where the first inequality is an application of Jensen's inequality,
the second inequality follows from Equation~\eqref{eq:help}
 the third inequality follows from Lipschitz continuity and the sixth inequality follows from another application of Jensen's inequality.
Consequently, by Lemma and the above chain of inequalities,
it follows that 
$\lim_{t\rightarrow \infty} \mathbf{E}[\|\nabla_{\mathbf{w}} \mathbb{F}(\mathbf{w}^t, \mathbf{v}^t)\|_2^2] = 0$.
Note that if $G$ is used, then by a similar argument, it follows
that $\lim_{t\rightarrow \infty} \mathbf{E}[\|\nabla_{\mathbf{v}} \mathbb{F}(\mathbf{w}^t, \mathbf{v}^t)\|_2^2] = 0$.
Consequently, if $G$ is used, then $\lim_{t\rightarrow \infty} \mathbf{E}[\|\nabla \mathbb{F}(\mathbf{w}^t, \mathbf{v}^t)\|_2^2] = 0$.
\end{proof}


\section{Proof of Proposition 1}

\begin{proposition}
Suppose $(\mathbf{x}, y)$ denotes a training sample and its corrupted label. For simplicity, let the MentorNet input $\phi(\mathbf{x}, y, \mathbf{w}) = \ell$ be the loss computed by the StudentNet model parameter $\mathbf{w}$. The MentorNet $g_m(\ell;\Theta) = v$, where $v$ is the sample weight. If $g_m$ decreases with respect to $\ell$, then there exists an underlying robust objective $F$:
\vspace{-2mm}
\[
F(\mathbf{w}) = \frac{1}{n} \sum_{i=1}^n \rho(\ell_i),
\vspace{-2mm}
\]
where $\rho(\ell_i)  = \int_0^{\ell_i} g_m(x;\Theta) d x$. In the special cases, $\rho(\ell)$ degenerates to the classical robust M-estimator: Huber~\cite{huber1964robust} and the log-sum
penalty~\cite{candes2008enhancing}.
\end{proposition}

\begin{proof}
Given a MentorNet $g_m$ and its fixed parameter $\Theta$. As $\phi(\mathbf{x}, y, \mathbf{w}) = \ell$, we have $g_m(\phi(\mathbf{x}, y, \mathbf{w}) ; \Theta) = g_m(\ell;\Theta)$.

$g_m$ is a neural network and hence is continuous with respect to $\ell$. Define $\rho(\ell)$ as 
\begin{equation}
\rho (\ell) = \int_0^{\ell}g_m(x;\Theta) d x
\end{equation}

Given the condition in the proposition, $g_m$ is decreasing with respect to $\ell$. The function $\rho$ is then bounded by its 1st term of the Taylor series about a point $\mathbf{w}^*$. We have:
\begin{equation}
\label{eq:taylor}
\rho(\phi(\mathbf{x}, y, \mathbf{w})) \le \rho(\phi(\mathbf{x}, y, \mathbf{w}^*)) + g_m(\phi(\mathbf{x}, y, \mathbf{w}^*);\Theta) (\phi(\mathbf{x}, y, \mathbf{w}) - \phi(\mathbf{x}, y, \mathbf{w}^*))
\end{equation}

According to~\cite{meng2015objective}, the right-hand side in Eq.~\eqref{eq:taylor} is a tractable surrogate for $\rho(\phi(\mathbf{x}, y, \mathbf{w}))$ and there exists an underlying robust objective. For the $i$-th sample, we have:
\begin{equation}
\label{eq:loss_rho}
\rho(\phi(\mathbf{x}_i, y_i, \mathbf{w})) = \rho(\ell_i) = \int_0^{\ell_i} g_m(x;\Theta) d x
\end{equation}

Finally, we have a robust objective derived from:
\begin{equation}
F(\mathbf{w}) = \frac{1}{n} \sum_{i=1}^n \rho(\phi(\mathbf{x}_i, y_i, \mathbf{w})) = \frac{1}{n} \sum_{i=1}^n \rho(\ell_i)
\end{equation}

Now we show the connection between Eq.~\eqref{eq:loss_rho} to the robust M-estimator. For simplicity, we assume that the loss $\ell \ge 0$ is non-negative for every training sample. For the Huber loss, there exists an $\Theta^*$ such that:
\begin{equation}
g_m(\ell; \Theta^*) = \begin{cases}
\frac{1}{2} & \ell \le \lambda^2 \\
\frac{\lambda}{2 \sqrt{\ell}} & \text{otherwise}
\end{cases},
\end{equation}
where $\lambda > 0$. It is easy to verify $g_m$ is decreasing with respect to $\ell$, and we have:
\begin{equation}
\begin{split}
\rho(\ell) = \int_0^{\ell} g_m(x;\Theta^*) d x = 
\begin{cases}
\frac{1}{2} \ell & \ell \le \lambda^2 \\
\lambda (\sqrt{\ell} -\frac{1}{2} \lambda) &\text{otherwise}
\end{cases}
\end{split}
\end{equation}

Therefore we have:
\begin{equation}
\label{eq:huber_loss}
\begin{split}
\rho(\ell^2) = 
\begin{cases}
\frac{1}{2} \ell^2 & \ell \le \lambda^2 \\
\lambda (\ell -\frac{1}{2} \lambda) &\text{otherwise}
\end{cases}
\end{split}
\end{equation}

Eq.~\eqref{eq:huber_loss} has a similar form of the Huber M-estimator~\cite{huber1964robust}.

Likewise, there exists an $\Theta^*$ and a positive $\epsilon$ such that
\begin{equation}
g_m(\ell; \Theta^*) = \frac{\lambda}{\ell + \epsilon} 
\end{equation}

Its underlying objective is identical to the log-sum penalty~\cite{candes2008enhancing}:
 
\begin{equation}
\rho(\ell) = \int_0^{\ell} g_m(x;\Theta^*) d x = \lambda \log(\ell +\epsilon) - \lambda \log(\epsilon) = \lambda \log( 1 + \frac{\ell}{\epsilon})
\end{equation}

It leads to the following log-sum penalty: 
\begin{equation}
F(\mathbf{w}) = \lambda \frac{1}{n} \sum_{i=1}^n  \log( 1 + \frac{\ell_i}{\epsilon})
\end{equation}
For the Lorentzian~\cite{black1996robust}. We have
\begin{equation}
g_m(\ell; \Theta^*) = \frac{2\ell}{2 \delta^2 + \ell^2}
\end{equation}

In special cases, we assume that $\delta$ is a small positive such that all sample loss $\ell \ge \sqrt{2} \delta$, the underlying objective is 
\begin{equation}
\rho(\ell) = \int_0^{\ell} g_m(x;\Theta^*) d x = \log(2 \delta^2 + \ell^2) - \log{2 \delta^2}= \log(1+ \frac{1}{2} (\frac{\ell}{\delta})^2)
\end{equation}
The above objective becomes Lorentzian~\cite{black1996robust}, also known as Lorentzian/Cauchy.

\end{proof}

\section{Comparison on MentorNet Architectures}
\label{sec:mentornet_arch}
This section examines \emph{MentorNet}'s architecture in terms of learning existing predefined curriculums (or weighting scheme). To generate the data for training MentorNet, we enumerate the feasible input space of $\mathbf{z}_i$ including the loss, difference to the loss moving average, label, and epoch percentage. For this experiment, the dataset contains a total of 300k samples. For each sample, we label a weight according to the weighting scheme in the predefined curriculum. For example, we compute the weight for focal loss by
\begin{equation}
\label{eq:focal_loss_weight}
v_i^* = [1-\exp\{-\ell_i\}]^\gamma,
\end{equation}
where $\gamma$ is a hyperparameter for smoothing the distribution. We consider the following predefined curriculums: self-paced~\cite{kumar2010self}, hard negative mining~\cite{felzenszwalb2010object}, linear weighting~\cite{jiang2015self}, focal loss~\cite{lin2017focal}, and prediction variance~\cite{chang2017active}. Besides, we also consider a temporal mixture weighting which is a mix of the self-paced (when the epoch percentage $<$ 50) and the hard negative mining (when the epoch percentage $\ge$ 50). In some curriculums, the form $G$ is unknown. We directly minimize the mean squared error between the MentorNet's output and the true weight.

We compare the MentorNet architecture in Fig. 1 in the paper to a simple logistic regression and 3 classical networks: a 2-layer Multiple Layer Perceptron (MLP), a 2-layer CNN with mean pooling, and an LSTM network (RNN) to sequentially encode the features of every example in a mini-batch. The same features are used across all networks. The objective is minimized by the Adam optimizer~\cite{kingma2015adam}. We use a very simple network for the proposed MentorNet. The bidirectional LSTM has a single layer of 10 base LSTM units (step size = 10). We use an embedding layer (size = 2) for the labels and an embedding layer (size = 5) for the integer epoch percentage between 0 and 99. The fully connected layer $fc_1$ uses the tangent activation function and has 20 hidden nodes.

\begin{figure*}[ht]
\vspace{-1mm}
\centering
\subfigure[Hard negative mining]
{
\includegraphics[width=0.4\linewidth,height=40mm]{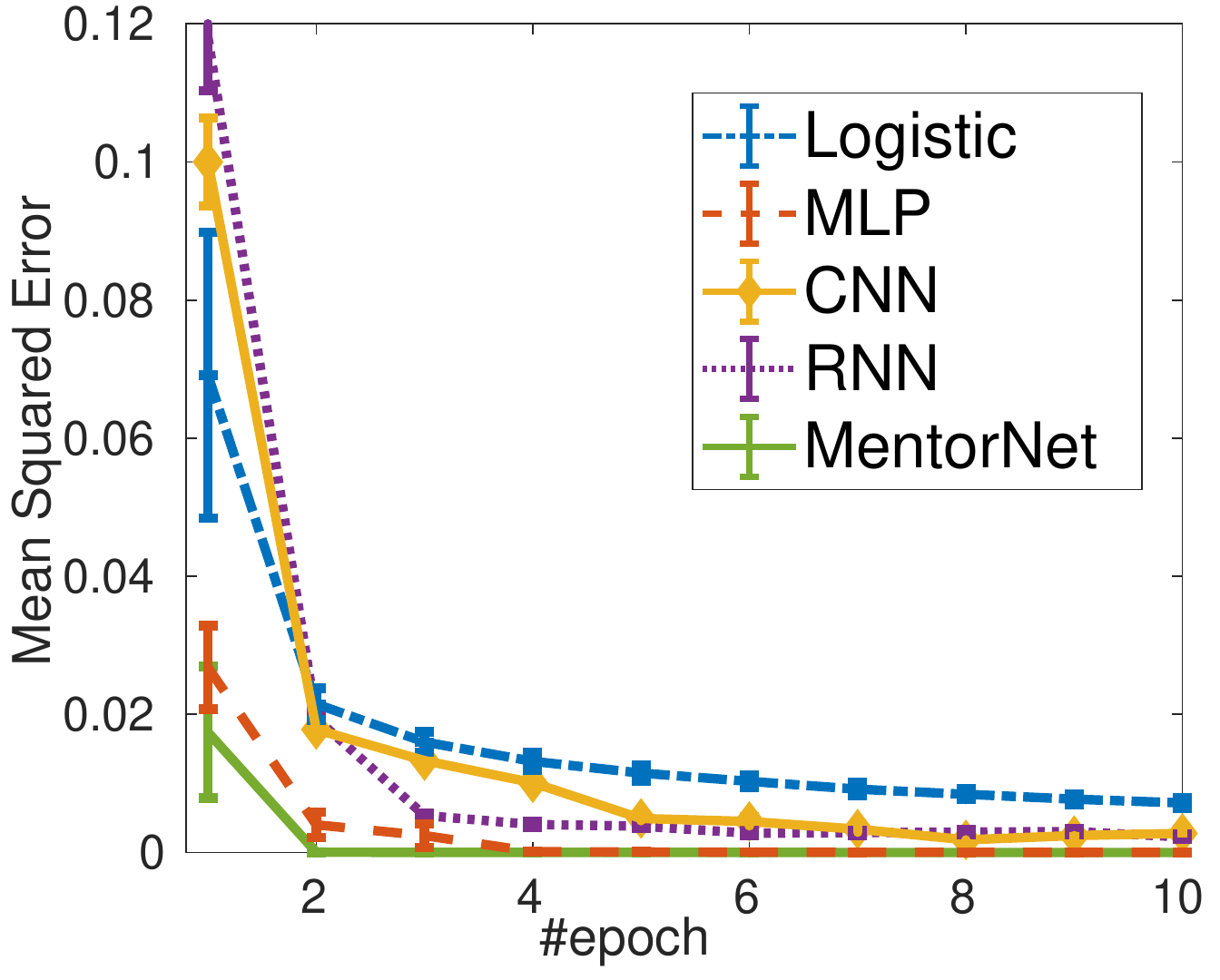}
}
\subfigure[Temporal mixture weighting]
{
\includegraphics[width=0.4\linewidth,height=40mm]{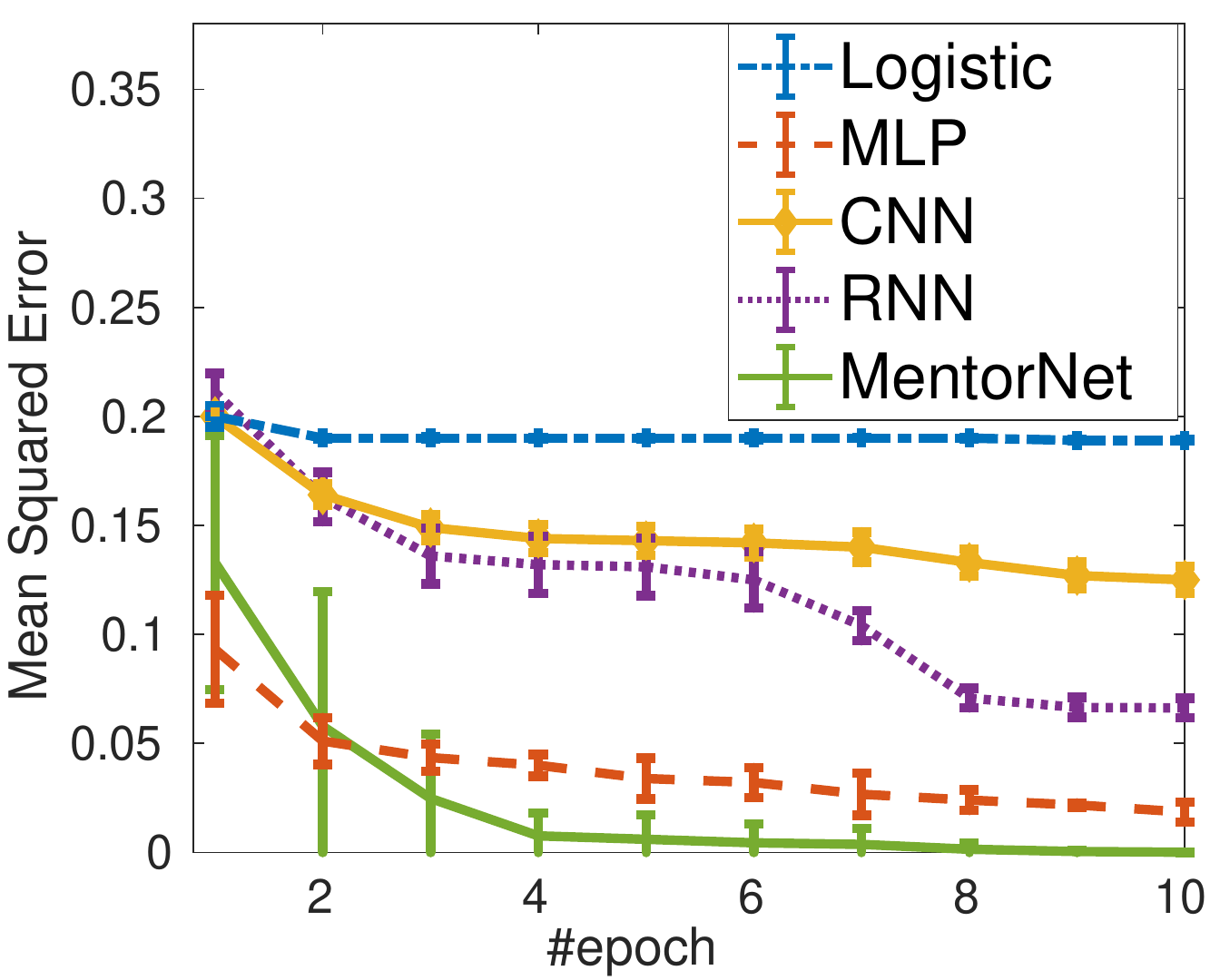}
}
\vspace{-2mm}
\caption{\label{fig:mentornet_convergence}Comparison of explicit and implicit MentorNet training.}
\end{figure*}

The performance is evaluated by the Mean Squared Error (MSE) to the true weight produced by each curriculum. Each experiment is repeated 5 times using random starting values, and the average MSE (with the 90\% confidence interval) is reported. Table~\ref{tab:mentornet_mse} shows the comparison results. As we see, the simple network structure MLP performs well except for complex weighting schemes that are prediction variance~\cite{chang2017active} and Temporal mixture weighting. Nevertheless, the bi-LSTM structure in Fig.1 of the paper performs better than other classical network architectures. Fig.~\ref{fig:mentornet_convergence} illustrates the error curve of different MentorNet architecture during training, where the $x$-axis is the training epoch and $y$-axis is the MSE.

\begin{table*}[ht]
\centering
\small
\caption{\label{tab:mentornet_mse}The MSE comparison of different MentorNet architecture on predefined curriculums.}
\begin{adjustbox}{max width=\textwidth}
\begin{tabular}{l||ccccc}
\hline
Weighting Scheme& Logistic & MLP & CNN & RNN & MentorNet (bi-LSTM)\\
\hline
Self-paced~\cite{kumar2010self}& 8.9$\pm$0.8E-3& 1.1$\pm$0.3E-5& 4.9$\pm$1.0E-2 & 1.6$\pm$1.0E-2 & \textbf{1.6$\pm$0.5E-6}\\
Hard negative mining~\cite{felzenszwalb2010object} &7.1$\pm$0.7E-3& 1.6$\pm$0.6E-5& 2.7$\pm$0.6E-3& 2.2$\pm$0.4E-3& \textbf{6.6$\pm$4.5E-7}\\
Linear weighting~\cite{jiang2015self} &9.2$\pm$0.1E-4& 1.2$\pm$0.4E-4& 1.1$\pm$0.2E-4& 2.0$\pm$0.3E-2& \textbf{4.4$\pm$1.3E-5}\\
Prediction variance~\cite{chang2017active}  &6.8$\pm$0.1E-3& 4.0$\pm$0.1E-3& 2.8$\pm$0.4E-2& 6.2$\pm$0.2E-3 & \textbf{1.4$\pm$0.7E-3}\\
Focal loss~\cite{lin2017focal} &1.7$\pm$0.0E-3& 6.0$\pm$3.5E-5 & 1.2$\pm$0.3E-2 & 1.2$\pm$0.3E-2 & \textbf{1.5$\pm$0.8E-5}\\
Temporal mixture weighting   &1.8$\pm$0.0E-1& 1.9$\pm$0.4E-2 & 1.2$\pm$0.6E-1 & 6.6$\pm$0.4E-2 & \textbf{1.2$\pm$1.1E-4}\\
\hline
\end{tabular}
\end{adjustbox}
\vspace{-2mm}
\end{table*}

\begin{figure*}[ht]
\vspace{-1mm}
\centering
\subfigure[Self-paced]
{
\includegraphics[width=0.4\linewidth]{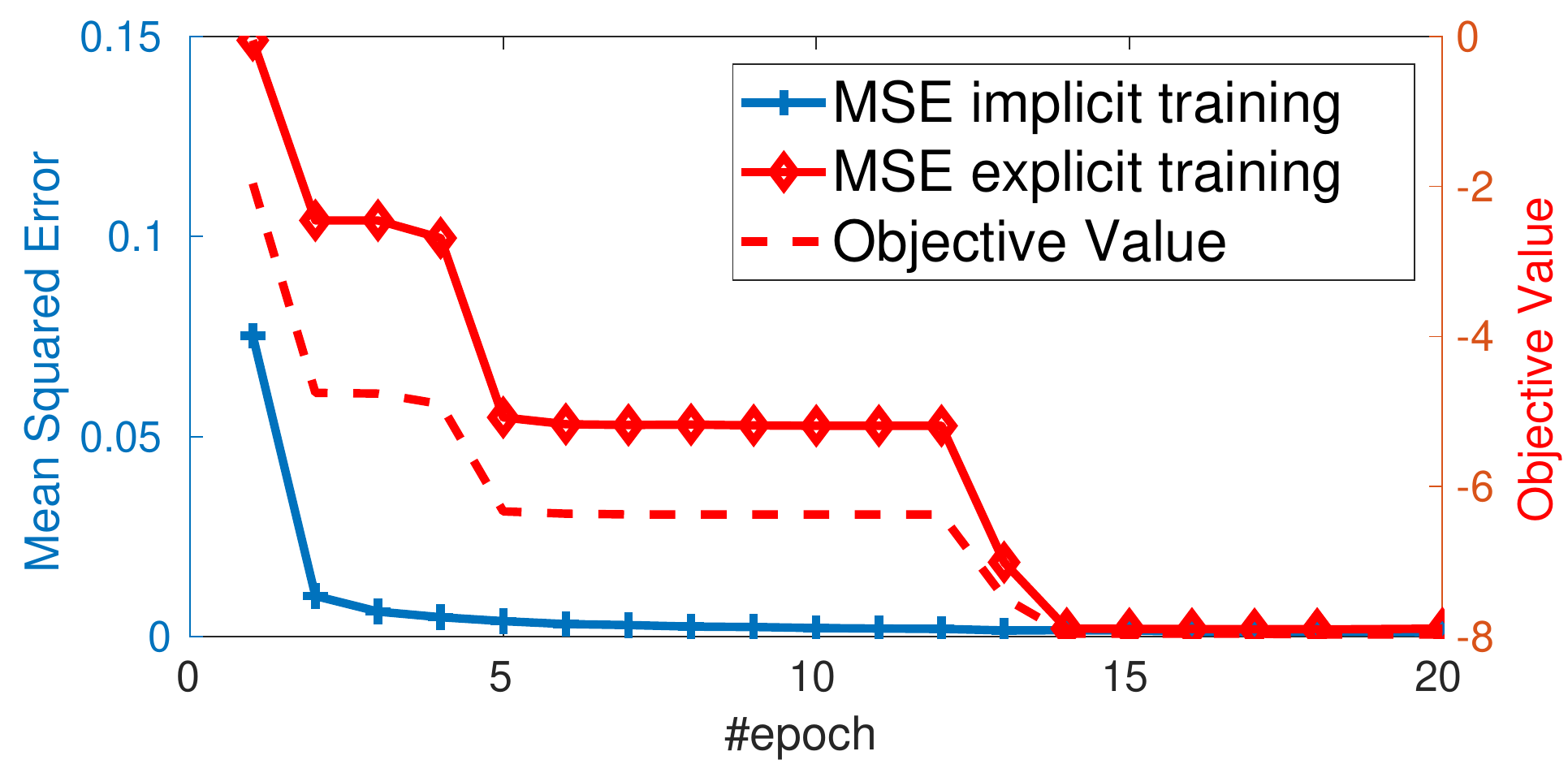}
}
\subfigure[Linear weighting]
{
\includegraphics[width=0.4\linewidth]{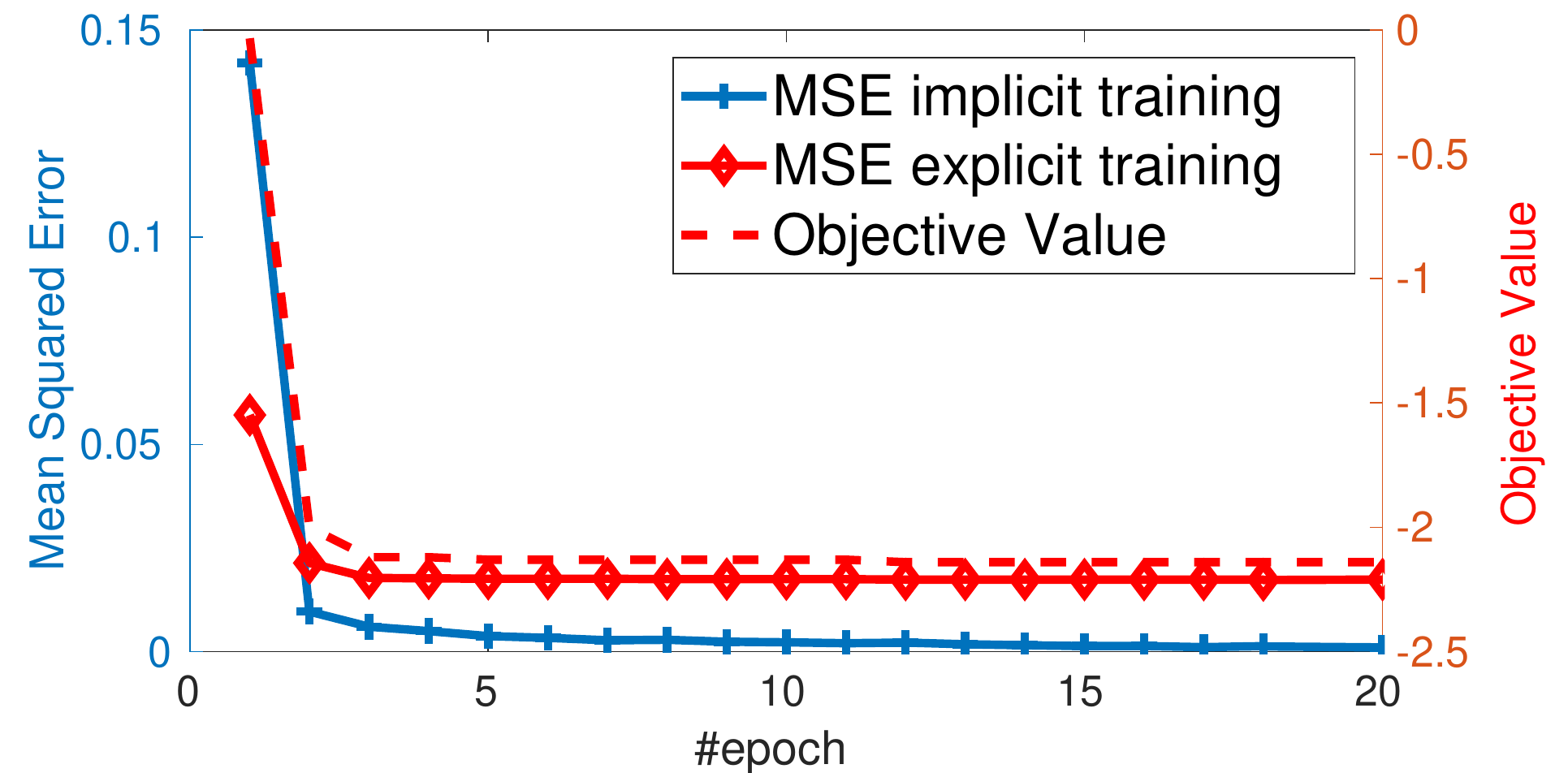}
}
\vspace{-2mm}
\caption{\label{fig:train_scheme_comp}Convergence comparison of different MentorNet architectures.}
\vspace{-5mm}
\end{figure*}

In Table~\ref{tab:mentornet_mse}, we learn a MentorNet by minimizing the MSE between its output and the true weight. We call it implicit training. When the form of $G$ is known, we can learn a MentorNet by directly minimizing Eq.(4) in the paper, called explicit training.
These two training approaches are theoretically identical and we empirically compare them on two curriculums of known $G$ in Fig.~\ref{fig:train_scheme_comp}. As shown, we found implicit training converges faster.

\section{Implementation Details}

\subsection{Dataset and StudentNet}\label{sec:data_student}
CIFAR-10 and CIFAR-100~\cite{krizhevsky2009learning} consist of 32 $\times$ 32 color images arranged in 10 and 100 classes. Both datasets contain 50,000 training and 10,000 validation images. The inception~\cite{szegedy2016rethinking} and wide-resnet-101~\cite{he2016deep,zagoruyko2016wide} are used as the StudentNet. Their implementations are based on the TensorFlow slim implementation\footnote{https://github.com/tensorflow/models/tree/master/research/slim}, where the inception is detailed in~\cite{zhang2017understanding} and the wider resnet is in~\cite{zagoruyko2016wide}. 

In training, the batch size is set to 128 and we train 39K iterations for resnet model and 120k for the inception model. The Step 12 in Algorithm 1 in the paper is implemented by the momentum SGD optimizer (momentum = 0.9) on a single GPU. We use the common learning rate scheduling strategy and set the starting learning rate as 0.1 and multiply it by a factor of 0.1 at the 19.5k, 25k and 30k iteration for the resnet model, and 78k for the inception model. Training in this way on the clean training dataset, the validation accuracy reaches about 81.4\% and 95.5\% for inception and resnet-101 on CIFAR-10, and about 49.2\% and 78.8\% on CIFAR-100.

By default, the StudentNet incorporates three types of regularization: 1) the weight decay which adds an $l_2$ norm of the model parameters into the learning objective; 2) data augmentation~\cite{krizhevsky2012imagenet} which augments the training images via domain-specific transformations random cropping, perturbation and contrast; 3) dropout~\cite{srivastava2014dropout} masks out network fully-connected layer outputs randomly. We use the best hyperparameter found on the clean training data. In the inception network, weight decay is set $4e{-3}$ and the dropout keep probability is 0.5. In the resnet-101, weight decay is $2e{-4}$ and the dropout keep probability is 1.0. In the Forgetting baseline~\cite{arpit2017closer}, we search the dropout parameter in the range of (0.2-0.9) and report the best accuracy on the validation set. The data augmentation is the same for the inception and the resnet-101: we pad 4 pixels to each side and randomly sampling a 32 x 32 crop, randomly flip the image horizontally (left to right), and then linearly scale the image to have zero mean and unit norm. Unless specified otherwise, the StudentNet of the same hyperparameter, discussed above, is used in all baseline and the proposed model.

ImageNet ILSVRC2012~\cite{deng2009imagenet} contains about 1.2 million training and 50k validation images, split into 1,000 classes. Each image is resized to 299 $\times$ 299 with 3 color channels. For the ImageNet, we use the inception-resnet v2 slim implementation\footnote{https://github.com/tensorflow/models/blob/master/research/slim/nets/inception\_resnet\_v2.py} as our StudentNet. We train the model on the ImageNet of 40\% noise. The Step 12 in Algorithm 1 in the paper is implemented as a distributed asynchronized momentum SGD optimizer (momentum = 0.9) on 50 GPUs. We set the batch size to 32 and train the model until it converges. That is 500 thousand steps for the StudentNet without any regularization and 1 million steps for the StudentNet with full regularization (weight decay, dropout and data augmentation). The starting learning rate is 0.05 and is decreased by a factor of 0.1 every 10 training epochs. The default dropout-keep-probability hyperparameter set to 0.8. In the Forgetting baseline, we set it to 0.2, which is the best parameter found on CIFAR-100. The weight decay is $4e{-5}$. The batch norm is used and its decay is set to 0.9997 and the epsilon is set to 0.001. The default data augmentation in the slim implementation is used. Training in this way on the clean training dataset, the validation accuracy is Hit@1=0.765.

\subsection{Baselines}
Regarding the baseline method. FullMode is the standard StudentNet trained using $l_2$ weight decay, dropout~\cite{srivastava2014dropout} and data augmentation~\cite{krizhevsky2012imagenet}. The same parameters discussed in Appendix~\ref{sec:data_student} are used. Forgetting was introduced in~\cite{arpit2017closer}. It is same as the FullModel except that the dropout parameter is tuned in the range of (0.2-0.9). For the self-paced learning~\cite{kumar2010self}, we gradually increase $\lambda$ in training by 20\%. Following~\cite{jiang2014self} we tune the parameter in the range of the 50th, 60th and 75th percentile of average sample loss. For the focal loss~\cite{lin2017focal}, we tune its $\gamma$ in the range of $\{1,2,3\}$ in our experiments. It is easy to verify that Eq.~\eqref{eq:focal_loss_weight} leads to the same classification objective in the focal loss~\cite{lin2017focal}, an award-winning method for object detection.  For the Reed~\cite{reed2014training}, we implement two versions: the soft and hard version. Let $q = [q_1, \dots, q_m]$ be the prediction (logits after softmax) for $m$ classes:

\begin{equation}
\ell_i = - (\beta \sum_{j=1}^m \mathds{1}(y_i = j) \log(q_j) + (1-\beta) \sum_{j=1}^m q_j \log(q_j)),
\end{equation}
where $\mathds{1}$ is the indicator function and a hard version:
\begin{equation}
\ell_i = - (\beta \sum_{j=1}^m \mathds{1}(y_i = j) \log(q_j) + (1-\beta) \max_{j} \log(q_j)),
\end{equation}
We tune the parameters $\beta$ in the range of $\{0.7, 0.8, 0.9, 0.95\}$.
Goldberger~\cite{goldberger2017training} is a recent baseline weakly-supervised learning method. We implement the S-Model in the paper by appending an additional layer to the StudentNet.

\subsection{Setups of Our Model}

The details about the MentorNet architecture (Fig. 1 in the paper) are discussed in Appendix~\ref{sec:mentornet_arch}. MentorNet PD is learned using the curriculum in Eq. (5) in the paper. The loss moving average $\ell_{pt}$ is set to the 75th-percentile loss in a mini-batch. We tune the hyperparameter $\lambda_1$ and $\lambda_2$. MentorNet DD is the learned \emph{data-driven} curriculum. It is trained on 5,000 images of true labels, randomly sampled from CIFAR-10. We learn MentorNet DD on this dataset and apply it to CIFAR-100 on which no true labels are used. We use the CIFAR-10 subset of the same level of the noise fraction corresponding to the CIFAR-100. CIFAR-10 and CIFAR-100 are two different datasets that have not only different classes but also the different number of classes. As CIFAR-100 and CIFAR-10 have the different number of classes, to apply a MentorNet, we fix the class label to 0.

Algorithm 1 is used to train the MentorNet together with the StudentNet. The decay factor in computing the loss moving average is set to 0.95. As mentioned in the paper, a burn-in process is used in the first 20\% training epoch for both MentorNet DD and MentorNet PD. We update and learn MentorNet twice after the learning rate is changed. That is on the 21\% and 75\% of the total epoch. More updates lead to insignificant performance difference. For each mini-batch, the weight decay parameter $\theta$ in Eq. (1) in the paper is normalized by the sum of the weight in a mini-batch. That is $\theta^t = \frac{1}{b} \theta^0 \sum_{i=1}^b \mathbf{v}_{\Xi_i}$, where $\theta^0$ is the original weight decay parameter. The same learning rate scheduling strategy in the StudentNet is used in Algorithm 1.


\bibliography{example_paper}
\bibliographystyle{icml2018}